\documentclass[journal]{IEEEtran}

\usepackage[pdftex]{graphicx}
\usepackage{cite}
\usepackage{amsopn}
\usepackage{booktabs} 
\usepackage{url}
\usepackage{color}
\usepackage{multirow}
\usepackage{xcolor}
\usepackage[switch]{lineno}
\usepackage{graphicx}
\usepackage{subfigure}
\usepackage{url}
\usepackage{epstopdf}
\usepackage{soul}
\usepackage{xcolor}
\usepackage{amsmath}
\usepackage{textcomp}
\usepackage{verbatim}
\usepackage{bm}
\usepackage{hyperref}
\usepackage{amssymb}

\begin{document}

\title{Physical-Virtual Collaboration Modeling for Intra-and Inter-Station Metro Ridership Prediction}

\author{Lingbo Liu, {\textit{Graduate Student Member, IEEE}},
        Jingwen Chen,
        Hefeng Wu, {\textit{Member, IEEE}},
        Jiajie Zhen, \\
        Guanbin Li, {\textit{Member, IEEE}},
        and~Liang Lin, {\textit{Senior Member, IEEE}}
\thanks{This work was supported in part by National Natural Science Foundation of China (NSFC) under Grant No. U1811463, 61876045 and 61976250, in part by the Natural Science Foundation of Guangdong Province under Grant No. 2017A030312006, in part by the Guangdong Basic and Applied Basic Research Foundation under Grant No. 2020B1515020048, in part by Zhujiang Science and Technology New Star Project of Guangzhou under Grant No. 201906010057 and was also sponsored by CCF-Tencent Open Research Fund. (\textit{Corresponding Author: Liang Lin.})}
\thanks{L. Liu, J. Chen, H. Wu, J. Zhen, G. Li, and L. Lin are with the School of Data and Computer Science, Sun Yat-Sen University, China, 510000 (e-mail: liulingb@mail2.sysu.edu.cn; chenjw93@mail2.sysu.edu.cn; wuhefeng@gmail.com; zhenjj@mail2.sysu.edu.cn; liguanbin@mail.sysu.edu.cn; linliang@ieee.org). Lingbo Liu and Jingwen Chen are co-first authors.}
}

\markboth{IEEE Transactions on Intelligent Transportation Systems}
{Liu \MakeLowercase{\textit{et al.}}: Physical-Virtual Collaboration Graph Network}

\maketitle

\begin{abstract}
Due to the widespread applications in real-world scenarios, metro ridership prediction is a crucial but challenging task in intelligent transportation systems. However, conventional methods either ignore the topological information of metro systems or directly learn on physical topology, and cannot fully explore the patterns of ridership evolution.
To address this problem, we model a metro system as graphs with various topologies and propose a unified Physical-Virtual Collaboration Graph Network (PVCGN), which can effectively learn the complex ridership patterns from the tailor-designed graphs. Specifically, a physical graph is directly built based on the realistic topology of the studied metro system, while a similarity graph and a correlation graph are built with virtual topologies under the guidance of the inter-station passenger flow similarity and correlation. These complementary graphs are incorporated into a Graph Convolution Gated Recurrent Unit (GC-GRU) for spatial-temporal representation learning. Further, a Fully-Connected Gated Recurrent Unit (FC-GRU) is also applied to capture the global evolution tendency. Finally, we develop a Seq2Seq model with GC-GRU and FC-GRU to forecast the future metro ridership sequentially.
Extensive experiments on two large-scale benchmarks (e.g., Shanghai Metro and Hangzhou Metro) well demonstrate the superiority of our PVCGN for station-level metro ridership prediction. Moreover, we apply the proposed PVCGN to address the online origin-destination (OD) ridership prediction and the experiment results show the universality of our method. Our code and benchmarks are available at {\color{blue}\url{https://github.com/HCPLab-SYSU/PVCGN}}.
\end{abstract}

\begin{IEEEkeywords}
Metro system, ridership prediction, graph convolutional networks, physical topology, virtual topology.
\end{IEEEkeywords}

\IEEEpeerreviewmaketitle
\section{Introduction}\label{sec:introduction}

\IEEEPARstart{M}{etro} is an efficient and economical travel mode in metropolises, and it plays an important role in the daily life of residents. By the end of 2018, 35 metro systems have been operated to serve tens of millions of passengers in Mainland China\footnote{\url{https://en.wikipedia.org/wiki/Urban_rail_transit_in_China}}. For instance, over 10 million metro trip transactions were made per day in 2018 for Beijing\footnote{\url{https://en.wikipedia.org/wiki/Beijing_Subway}} and Shanghai\footnote{\url{https://en.wikipedia.org/wiki/Shanghai_Metro}}. Such huge metro ridership poses great challenges for urban transportation and any carelessness of traffic management may result in citywide congestions.
For improving the service efficiencies of metro systems, a fundamental problem is how to accurately forecast the ridership (e.g., inflow and outflow) of each station, which is termed as station-level metro ridership prediction in this work. Due to its potential applications in traffic dispatch and route planning, this problem has become a hotspot research topic~\cite{li2017forecasting,tang2018forecasting,gong2018network,chen2019subway,fang2019gstnet,hao2019sequence} in the community of intelligent transportation systems (ITSs).

Over the past decade, massive efforts have been made to address the traffic states (e.g., flow, speed and demand) prediction. In early works~\cite{lippi2013short,guo2014adaptive,kumar2015short}, the raw data of traffic states at each time interval was usually transformed to be a vector/sequence and the time series models~\cite{williams1998urban,guo2014adaptive} were applied for prediction. However, this data format failed to maintain the spatial information of locations and the topological connection information between two locations.
In recent years, deep neural networks (e.g., Long Short-term Memory~\cite{xingjian2015convolutional} and Gated Recurrent Unit~\cite{chung2014empirical}) have been widely used for citywide traffic prediction~\cite{zhang2017deep,yao2018deep,liang2018geoman,liu2019contextualized,yao2019revisiting,zhang2019flow,liu20120dynamic}.
These works usually partitioned the studied cities into regular grid maps on the basis of geographical coordinate and organized the collected traffic state data as Euclidean 2D or 3D tensors, which can be straightway fed into convolutional networks for automatic representation learning. Nevertheless, this manner is unsuitable for metro systems, since their topologies are irregular graphs and their data structures are non-Euclidean.
Although the transaction records of a metro system can be rendered as a grid map~\cite{ma2019parallel}, it is inefficient to learn ridership evolution patterns from the rendered map, which is very sparse and can not maintain the connection information of two stations.

In general, the challenges of metro ridership prediction lie in how to efficiently model the non-Euclidean structures of metro systems and fully capture the ridership evolution patterns. Although the emerging Graph Convolution Networks (GCN~\cite{bruna2014spectral,duvenaud2015convolutional,kipf2016semi}) have been proven to be general for non-Euclidean data embedding, how to construct the reasonable graphs in GCN still is an open problem and the construction strategy is varying in different tasks~\cite{li2017scene,wang2018describe,zhong2019graph,chen2019learning}. Some recent works \cite{li2018diffusion,yu2018spatio,cui2018traffic,geng2019spatiotemporal,zhao2019t,xiao2020demand,zheng2020gman} have applied GCN to traffic prediction and most of them directly build geographical graphs based on the physical topologies of the studied traffic systems. However, this simple strategy is suboptimal for metro ridership prediction, since it only learns the local spatial dependency of neighboring stations and can not fully capture the inter-station flow patterns in a metro system. Therefore, except for the physical topologies, we should construct some more reasonable graphs with human domain knowledge, such as:
\begin{itemize}
\item {\bf{Inter-station Flow Similarity:}}
Intuitively, two metro stations in different regions may have similar evolution patterns of passenger flow, if their located regions share the same functionality (e.g., office districts). Even though these stations are not directly linked in the real-world metro system, we can connect them in GCN with a virtual edge to jointly learn the evolution patterns.
\item {\bf{Inter-station Flow Correlation:}}
In general, the ridership between every two stations is not uniform and the direction of passenger flow implicitly represents the correlation of two stations. For instance, if (i) the majority of inflow of station $a$ streams to station $b$, or (ii) the outflow of station $a$ primarily comes from station $b$, we argue that the stations $a$ and $b$ are highly correlated. Under such circumstances, these stations could also be connected to learn the ridership interaction among stations.
\end{itemize}

Based on the above observations, we propose a unified Physical-Virtual Collaboration Graph Network (PVCGN) to predict the future metro ridership in an end-to-end manner. To fully explore the ridership evolution patterns, we utilize the metro physical topology information and human domain knowledge to construct three complementary graphs.
First, a physical graph is directly formed on the basis of the realistic topology of the studied metro system. Then, a similarity graph and a correlation graph are built with virtual topologies respectively based on the passenger flow similarity and correlation among different stations. In particular, the similarity score of two stations is obtained by computing the warping distance between their historical flow series with Dynamic Time Warping (DTW~\cite{berndt1994using}), while the correlation ratio is determined by the historical origin-destination distribution of ridership.
These tailor-designed graphs are incorporated into an extended Graph Convolution Gated Recurrent Unit to collaboratively capture the ridership evolution patterns. Furthermore, a Fully-Connected Gated Recurrent Unit is utilized to learn the semantic feature of global evolution tendency. Finally, we apply a Seq2Seq model~\cite{sutskever2014sequence} to sequentially forecast the metro ridership at the next several time intervals.
To verify the effectiveness of our PVCGN, we conduct extensive experiments on two large-scale benchmarks (i.e., Shanghai Metro and Hangzhou Metro) and the evaluation results show that our approach outperforms existing state-of-the-art methods under various comparison circumstances. For verifying the universality of our method, we further employ the proposed PVCGN to forecast the online origin-destination (OD) ridership and the experiment results also demonstrate the effectiveness of PVCGN for OD ridership prediction.

In summary, our major contributions are four-fold:
\begin{itemize}
\item We develop a unified Physical-Virtual Collaboration Graph Network (PVCGN) to address the station-level metro ridership prediction. Specifically, PVCGN incorporates a physical graph, a similarity graph and a correlation graph into a Graph Convolution Gated Recurrent Unit to facilitate the spatial-temporal representation learning.
\item The physical graph is built based on the realistic topology of a metro system, while the other two virtual graphs are constructed with human domain knowledge to fully exploit the ridership evolution patterns.
\item Extensive experiments on two real-world metro ridership benchmarks show that our PVCGN comprehensively outperforms state-of-the-art methods for station-level ridership prediction.
\item As a general model, our PVCGN can be directly employed for online origin-destination ridership prediction and also achieves superior performance.
\end{itemize}

The remaining parts of this paper are organized as follows. We first investigate the deep learning on graphs and some related works of traffic states prediction in Section~\ref{sec:review}. The proposed PVCGN is then introduced systematically in Section~\ref{sec:method}. We conduct extensive comparisons for station-level metro ridership prediction in Section~\ref{sec:experiment} and extend PVCGN to forecast online OD ridership in Section~\ref{sec:OD}. Finally, we conclude this paper and discuss future works in Section~\ref{sec:conclusion}.

\section{Related Work}\label{sec:review}
\subsection{Deep Learning on Graphs}
In machine learning, Euclidean data refers to the signals with an underlying Euclidean structure~\cite{liu2020cp,liu2018crowd,liu2019facial,liu2019crowd,chen2020knowledge,liu2020efficient
,yan2019semi} (such as speeches, images, and videos). Although deep Convolutional/Recurrent Neural Networks (CNN/RNN) can handle Euclidean data successfully, it is still challenging to deal with non-Euclidean data (e.g., graphs), which is the data structure of many applications. To address this issue, Graph Convolution Networks (GCN) have been proposed to automatically learn feature representation on graphs. For instance, Brunaat et al.~\cite{bruna2014spectral} introduced a graph-Laplacian spectral filter to generalize the convolution operators in non-Euclidean domains. Defferrard et al.~\cite{defferrard2016convolutional} presented a formulation of CNN with spectral graph theory and designed fast localized convolutional filters on graphs. Atwood and Towsley~\cite{atwood2016diffusion} developed a spatial-based graph convolution and regarded it as a diffusion process, in which the information of a node was transferred to its neighboring nodes with a certain transition probability. Velivckovic et al.~\cite{velivckovic2017graph} assumed the contributions of neighboring nodes to the central node were neither identical, thus proposed a Graph Attention Network. Wu et al.~\cite{wu2019simplifying} reduced the complexity of GCN through successively removing nonlinearities and collapsing weight matrices between consecutive layers. Seo et al.~\cite{seo2018structured} incorporated graph convolution and RNN to simultaneously exploit the graph spatial and dynamic information for structured sequences learning.

Recently, GCN has been widely applied to address various tasks and the graph construction strategy varied in different works. For instance, in computer vision, Jiang et al.~\cite{jiang2018hybrid} utilized the co-occurrence probability, attribute correlation and spatial correlation of objects to build three graphs for large-scale object detection. Chen et al.~\cite{chen2019learning} constructed a semantic-specific graph based on the statistical label co-occurrence for multi-label image recognition.
In natural language processing, Beck et al.~\cite{beck2018graph} used source dependency information to built a Levi graph~\cite{levi1942finite} for neural machine translation.
For semi-supervised document classification, Kipf and Welling~\cite{kipf2016semi} introduces a first-order approximation of spectral graph~\cite{defferrard2016convolutional} and constructed their graphs based on citation links.
In data mining, the relations between items-and-items, users-and-users and users-and-items were usually leveraged to construct graph-based recommender systems~\cite{ying2018graph}. In summary, how to build a graph is an open problem and we should flexibly design the topology of a graph for a specific task.

\subsection{Traffic States Prediction}
Accurately forecasting the future traffic states is crucial for intelligent transportation systems and numerous models have been proposed to address this task~\cite{bolshinsky2012traffic,barros2015short,nagy2018survey}.
In early works~\cite{li2012prediction,lippi2013short,guo2014adaptive,kumar2015short,dell2015time}, mass traffic data was collected from some specific locations and the raw data at each time interval was arranged as a vector (sequence) in a certain order. These vectors were further fed into time series models for prediction. A representative work was the data aggregation (DA) model~\cite{tan2009aggregation}, in which the moving average (MA), exponential smoothing (ES) and autoregressive MA (ARIMA) were simultaneously applied to forecast traffic flow. However, this simple data format was inefficient due to the lack of spatial information, and these basic time series models failed to learn the complex traffic patterns. Therefore, the above-mentioned works were far from satisfactory in complex traffic scenarios.

In recent years, deep neural networks have become the mainstream approach in this field. For instance, Zhang et al.~\cite{zhang2017deep} utilized three residual networks to learn the closeness, period and trend properties of crowd flow.
Wang et al.~\cite{wang2017deepsd} developed an end-to-end convolutional neural network to automatically discover the supply-demand patterns from the car-hailing service data. Zhang et al.~\cite{zhang2019flow} simultaneously predicted the region-based flow and inter-region transitions with a deep multitask framework. Subsequently, RNN and its various variants are also widely adopted to learn the temporal patterns. For instance, Yao et al.~\cite{yao2018deep} proposed a Deep Multi-View Spatial-Temporal Network for taxi demand prediction, which learned the spatial relations and the temporal correlations with deep CNN and Long Short-Term Memory (LSTM~\cite{xingjian2015convolutional}) unit respectively. Liu et al.~\cite{liu2018attentive} developed an attentive convolutional LSTM network to dynamically learn the spatial-temporal representations with an attention mechanism. In \cite{yao2019revisiting}, a periodically shifted attention mechanism based LSTM was introduced to capture the long-term periodic dependency and temporal shifting. To fit the required input format of CNN and RNN, most of these works divided the studied cities into regular grid maps and transformed the raw traffic data to be tensors. However, this preprocessing manner is ineffective to handle the traffic systems with irregular topologies, such as metro systems and road networks.

To improve the generality of the above-mentioned methods, some researchers have attempted to address this task with Graph Convolutional Networks. For instance, Li et al.~\cite{li2018diffusion} modeled the traffic flow as a diffusion process on a directed graph and captured the spatial dependency with bidirectional random walks, while Zhao et al.~\cite{zhao2019t} proposed a temporal graph convolutional network for traffic forecasting based on urban road networks.
Guo et al.~\cite{guo2019attention} and Zheng et al.~\cite{zheng2020gman} introduced attention mechanisms into spatial-temporal graph networks to dynamically model the impact of various factors for traffic prediction. Wu et al.~\cite{wu2019graph} developed an adaptive dependency matrix with node embedding to precisely capture the hidden spatial dependency. Bai et al.~\cite{bai2019stg2seq} utilized a hierarchical graph convolutional structure to capture both spatial and temporal correlations for multi-step passenger demand prediction. Song et al.~\cite{song2020spatial} developed a Spatial-Temporal
Synchronous Graph Convolutional Networks (STSGCN), which captured the complex localized spatial-temporal correlations through a spatial-temporal synchronous modeling mechanism. Recently, GCN has also been employed to metro ridership prediction. In \cite{han2019predicting}, graph convolution operations were applied to capture the irregular spatiotemporal dependencies along with the metro network, but their graph was directly built based on the physical topology of metro systems. In constant, we combine the physical topologies information and human domain knowledge to construct three collaborative graphs with various topologies, which can effectively capture the complex patterns.

The most relevant work to ours is ST-MGCN~\cite{geng2019spatiotemporal}, which incorporated a neighborhood graph (NGraph), a transportation connectivity graph (TGraph), and a functional similarity graph (FGraph) for ride-hailing demand prediction.  The differences between our PVCGN and ST-MGCN are two-fold.
{\bf{First}}, ST-MGCN relied heavily on the extra information of road networks (e.g., motorway and highway) and Point of Interests (POI) for graph construction. However, this information is inaccessible in many scenarios. In contrast, our PVCGN does not require any external information, and our graphs can be directly built with the spatial topology information and the historical ridership data. Thus our method is more flexible and universal for traffic prediction.
{\bf{Second}}, ST-MGCN paid more attention to building the physical graphs (i.e., NGraph and TGraph) based on real-world topologies and only built a virtual graph (i.e., FGraph) with the external POI information. In contrast, except for the physical graph, our PVCGN fully explores the potential traffic patterns (such as inter-station flow similarity and OD correlation) for virtual graph construction. Therefore, our method can learn more comprehensive and knowledgeable representation for traffic prediction.

\begin{figure*}
    \centering
    \includegraphics[width=1.95\columnwidth]{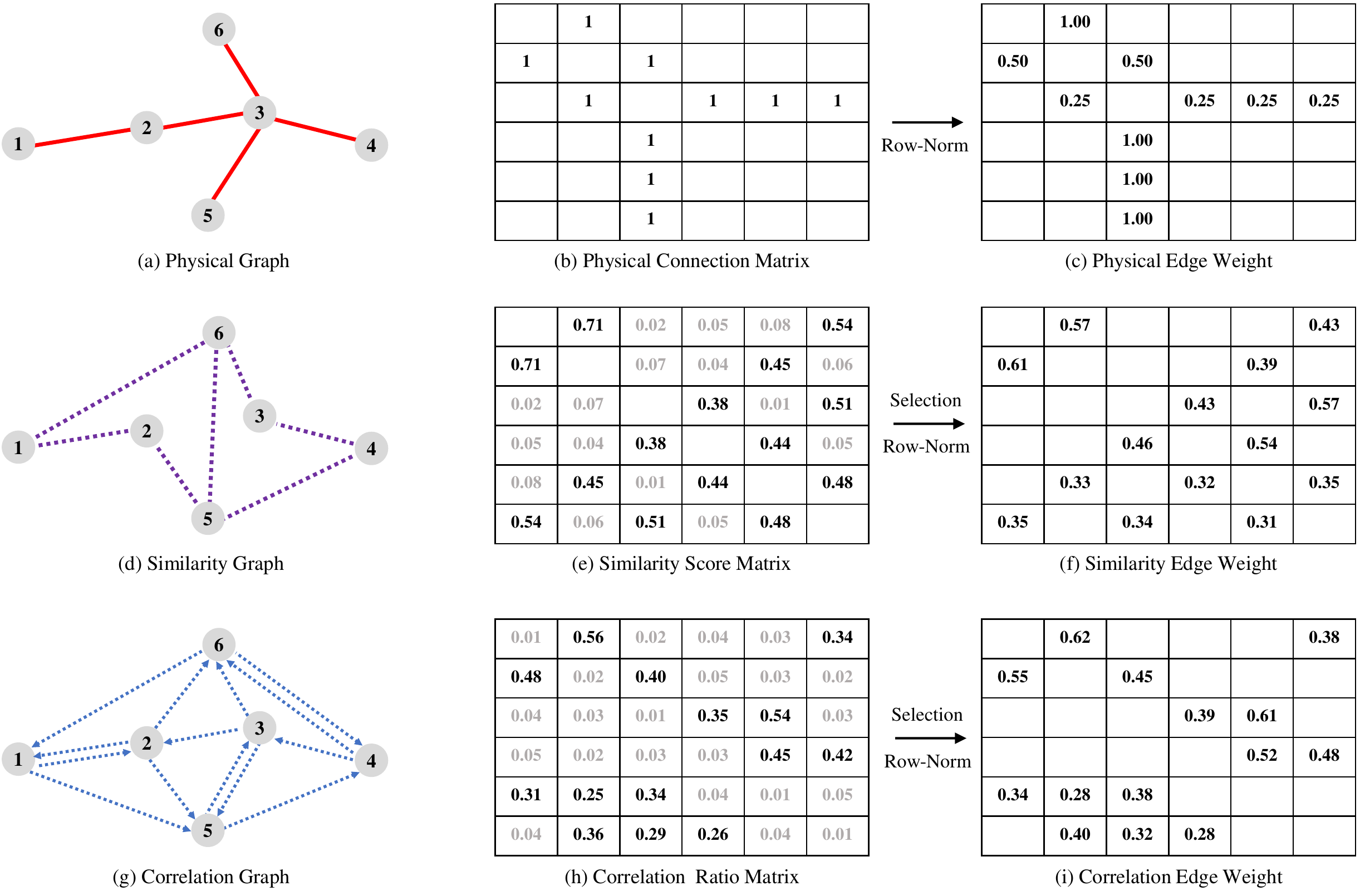}
    \vspace{-2mm}
    \caption{Illustration of the proposed physical-virtual graphs. In this figure, we take a metro system with six stations as an example to illustrate the construction strategy of our physical-virtual graphs.
    {\bf{First Row:}} (a) is the physical graph built based on the physical topology of the studied metro system. We perform row normalization on the physical connection matrix (b) to obtain the edge weights (c).
    {\bf{Second Row:}} (d) is the built similarity graph and its edges are determined by the station pairs with high similarity scores in the matrix (e). The edge weights (d) are computed by conducting row normalization on the similarity scores of the selected station pairs.
    {\bf{Third Row:}} (g) is the built correlation graph. We compute the correlation ratio matrix (e) by measuring the origin-destination distribution of ridership and select the station pairs with high correlation ratios to construct edges. The edge weights (i) are obtained by conducting row normalization on the correlation ratios of the selected station pairs. {\textit{Note that the origin is the top left corner for all matrices.}}
    }
    \label{fig:graphs}
    \vspace{0mm}
\end{figure*}

\subsection{Traffic Origin-Destination Prediction}
Traffic origin-destination (OD) prediction is a challenging task, which aims to forecast the traffic flow or demand between any two positions. Some early works~\cite{zhou2007structural,hazelton2001inference,djukic2014dynamic} usually employed time series models (e.g., Kalman filter) to estimate the OD flow, while recent works developed various deep neural networks to forecast the OD matrices.
For instance, Liu et al.~\cite{liu2019contextualized} proposed a Contextualized Spatial-Temporal Network that incorporated local spatial
context, temporal evolution context, and global correlation context to forecast taxi OD demand. Chu et al.~\cite{chu2019deep} developed a Multi-Scale Convolutional LSTM Network for taxi OD flow prediction. Wang et al.~\cite{wang2019origin} developed a Grid-Embedding based Multi-task Learning framework that applied graph convolutions among geographical and semantic neighbors to model the OD transferring patterns. Shi et al.~\cite{shi2020predicting} utilized long short-term memory units to extract temporal features for each OD pair and then learned the spatial dependency of origins and destinations by a two-dimensional graph convolutional network.
In the aforementioned ride-hailing applications, the origin and destination of a passenger are known once a taxi request is generated. However, in online metro systems, the destination of a passenger is unknown until it reaches the destination station, so we can not obtain the complete OD distribution immediately to forecast the future OD demand. To address this issue, Gong et al.~\cite{gong2020online} used some indication matrices to mask and neglect the potential unfinished metro orders. In our work, we apply the proposed PVCGN to handle this task by learning a mapping from the historical incomplete OD demands to the future complete OD demands, and more details can be found in Section~\ref{sec:OD}.

\section{Methodology}\label{sec:method}
In this work, we propose a novel Physical-Virtual Collaboration Graph Network (PVCGN) for station-level metro ridership prediction. Based on the physical topology of a metro system and human domain knowledge, we construct a physical graph, a similarity graph and a correlation graph, which are incorporated into a Graph Convolutional Gated Recurrent Unit (GC-GRU) for local spatial-temporal representation learning. Then, a Fully-Connected Gated Recurrent Unit (FC-GRU) is applied to learn the global evolution feature. Finally, we develop a Seq2Seq framework with GC-GRU and FC-GRU to forecast the ridership of each metro station. %

We first define some notations of ridership prediction before introducing the details of PVCGN. The ridership data of station $i$ at time interval $t$ is denoted as $\bm{X}_t^i \in \mathbb{R}^2$, where these two values are the passenger counts of inflow/outflow. The ridership of the whole metro system is represented as a signal $\bm{X}_{t} = (\bm{X}_t^1,\bm{X}_t^2,...,\bm{X}_t^N ) \in \mathbb{R}^{2 \times N}$, where $N$ is the number of stations. Given a historical ridership sequence, our goal is to predict a future ridership sequence:
\begin{equation}
\begin{small}
\hat{\bm{X}}_{t+1},\hat{\bm{X}}_{t+2},..., \hat{\bm{X}}_{t+m}=\text{PVCGN}(\bm{X}_{t-n+1},\bm{X}_{t-n+2},...,\bm{X}_{t})
\end{small}
\end{equation}%
where $n$ refers to the length of the input sequence and $m$ is the length of the predicted sequence. For the convenience in following subsections, we also denote the whole historical ridership of station $i$ as a vector $\bm{X}^i \in \mathbb{R}^{2T}$, where $T$ is the number of time intervals in a training set.

\subsection{Physical-Virtual Graphs}
In this section, we describe how to construct the physical graph and two virtual graphs. By definition, a graph is composed of nodes, edges as well as the weights of edges. In our work, the physical graph, similarity graph and correlation graph are denoted as $\mathcal{G}_p=(\mathcal{V},\mathcal{E}_p, {W}_p)$, $\mathcal{G}_s=(\mathcal{V},\mathcal{E}_s, {W}_s)$ and $\mathcal{G}_c=(\mathcal{V},\mathcal{E}_c, {W}_c)$, respectively.
$\mathcal{V}$ is the set of nodes ($|\mathcal{V}|=N$) and each node represents a real-world metro station. Note that these three graphs share the same nodes, but have different edges and edge weights. $\mathcal{E}_p$, $\mathcal{E}_s$ and $\mathcal{E}_c$ are the edge sets of different graphs. For a specific graph $\mathcal{G}_\alpha$  ($\alpha=p,s,c$), $W_\alpha \in \mathbb{R}^{N{\times}N} $ denotes the weights of all edges. Specifically, $W_\alpha(i,j)$ is the weight of an edge from node $j$ to node $i$.

\subsubsection{\bf{Physical Graph}}
$\mathcal{G}_p$ is directly built based on the physical topology of the studied metro system.
An edge is formed to connect node $i$ and $j$ in $\mathcal{E}_p$, if the corresponding station $i$ and $j$ are connected in real world.
To calculate the weights of these edges, we first construct a physical connection matrix $P \in \mathbb{R}^{N{\times}N}$. As shown in Fig.\ref{fig:graphs}-(a,b), $P(i,j)=1$ if there exists an edge between node $i$ and $j$, or else $P(i,j)=0$. To avoid the repetitive computation of graph self-loop, each diagonal value $P(i,i)$ is directly set to 0 and the self-loop would be uniformly computed once for multi-graphs in Eq.\ref{eq:gc}. Finally, the edge weight ${W}_p$ is obtained by performing a linear normalization on each row (See Fig.\ref{fig:graphs}-c). Specifically, ${W}_p(i,j)$ is computed by:
\begin{equation}
    {W}_p(i,j) = \frac{P(i,j)}{\sum_{k=1}^N P(i,k)}
\end{equation}

\subsubsection{\bf{Similarity Graph}}\label{sec:sg}
In this section, the similarities of metro stations are used to guide the construction of $\mathcal{G}_s$. First, we construct a similarity score matrix $S \in \mathbb{R}^{N{\times}N}$ by calculating the passenger flow similarities between every two stations. Specifically, the score $S(i,j)$ between station $i$ and $j$ is computed with Dynamic Time Warping (DTW~\cite{berndt1994using}):
\begin{equation}
    S(i,j) = exp(-\text{DWT}(\bm{X}^i, \bm{X}^j)),
\end{equation}%
where DTW is a general algorithm for measuring the distance between two temporal sequences. Note that $S(i,i)$ is also directly set to 0. Based on the matrix $S$, we select some station pairs to build edges $\mathcal{E}_s$. The selection strategy is flexible. For instance, these virtual edges can be determined with a predefined similarity threshold, or be built by choosing the top-$k$ station pairs with high similarity scores. More selection details can be found in Section~\ref{sec:implementation}. Finally, we calculate the edge weights ${W}_s$ by conducting row normalization on $S$:
\begin{equation}
    {W}_s(i,j) = \frac{S(i,j)}{\sum_{k=1}^N S(i,k) \cdot L(\mathcal{E}_s, i,k)}
\end{equation}%
where $L(\mathcal{E}_s, i,k)=1$ if $\mathcal{E}_s$ contains an edge connecting node $i$ and $k$, or else $L(\mathcal{E}_s, i,k)=0$. A toy example of similarity graph is shown in Fig.\ref{fig:graphs}-(d,e,f) and we can observe that matrix $S$ is symmetrical, but matrix ${W}_s$ is asymmetrical due to the row normalization.

\subsubsection{\bf{Correlation Graph}}
We utilize the origin-destination distribution of ridership to build the virtual graph $\mathcal{G}_c$. First, we construct a correlation ratio matrix $C \in \mathbb{R}^{N{\times}N}$. Specifically, $R(i,j)$ is computed by:
\begin{equation}
    {C}(i,j) = \frac{D(i,j)}{\sum_{k=1}^N D(i,k)}
\end{equation}%
where $D(i,j)$ is the total number of passengers that traveled from station $j$ to station $i$ in the whole training set. Note that ${C}(i,i)$ is computed, since there are a small number of passengers that entered and exited at the same station in the real world. We use the similar selection strategy described in Section~\ref{sec:sg} to select some station pairs for edge construction. Finally, the edge weights ${W}_c$ is calculated by:
\begin{equation}
    {W}_c(i,j) = \frac{C(i,j)}{\sum_{k=1}^N C(i,k) \cdot L(\mathcal{E}_c, i,k)}
\end{equation}
One example of correlation graph is shown in Fig.\ref{fig:graphs}-(d,e,f) and we can see that $\mathcal{G}_c$ is a directed graph, since ${R}(i,j) \not= {R}(j,i)$.

\begin{figure*}
    \centering
    \includegraphics[width=1.45\columnwidth]{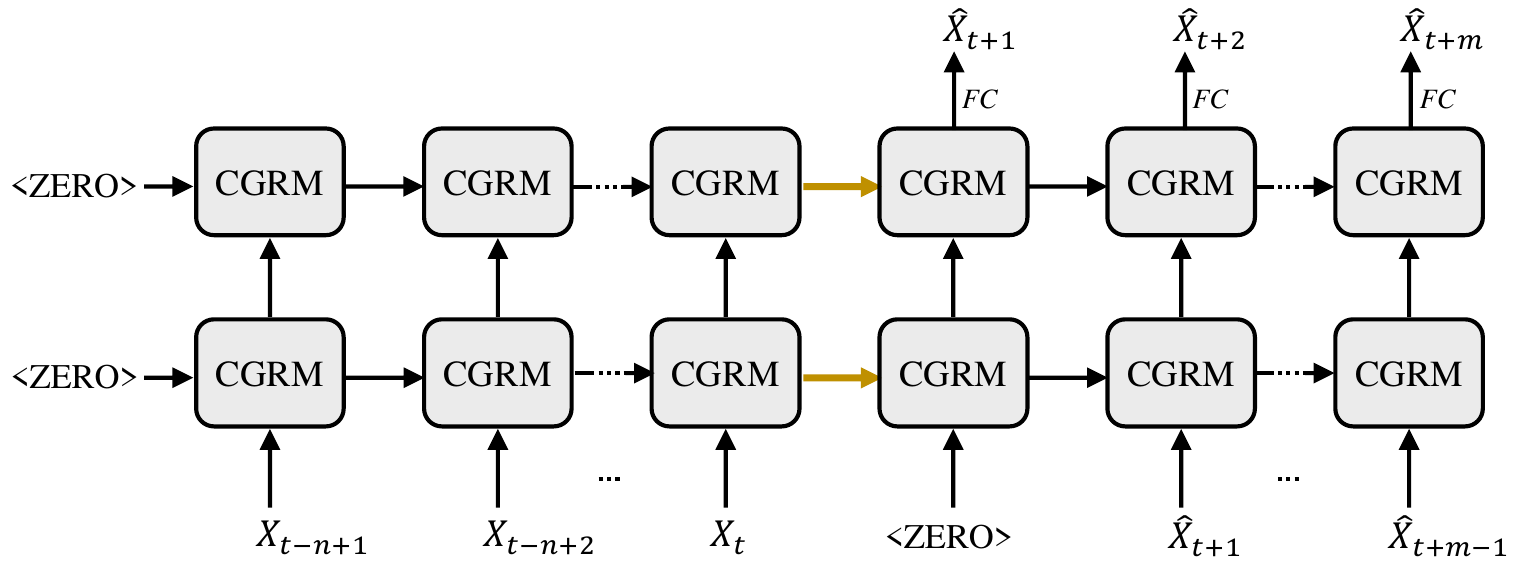}
    \vspace{-2mm}
    \caption{Overview of Physical-Virtual Collaboration Graph Network (PVCGN) for station-level metro ridership prediction. PVCGN consists of an encoder and a decoder, both of which contains two Collaborative Gated Recurrent Modules (CGRMs). The encoder takes $\{\bm{X}_{t-n+1},\bm{X}_{t-n+2},...,\bm{X}_{t}\}$ as input and the decoder forecasts a future ridership sequence $\{\hat{\bm{X}}_{t+1},\hat{\bm{X}}_{t+2},..., \hat{\bm{X}}_{t+m}\}$ with fully-connected (FC) layers.}
    \label{fig:arch}
\end{figure*}

\subsection{Graph Convolution Gated Recurrent Unit}
As an alternative of LSTM~\cite{xingjian2015convolutional}, Gated Recurrent Unit has been widely used for temporal modeling and it was usually implemented with standard convolution or full-connection.
In this section, we incorporate the proposed physical-virtual graphs to develop a unified Graph Convolution Gated Recurrent Unit (GC-GRU) for spatial-temporal feature learning.

We first formulate the convolution on the proposed physical-virtual graphs.
Let us assume that the input of graph convolution is $I_t = \{I_t^1,I_t^2,...,I_t^N\}$, where $I_t^i$ can be the ridership data $\bm{X}_t^i$ or its feature. The parameters of this graph convolution are denoted as $\Theta$. By definition of convolution, the output feature $f(I_t^i) \in \mathbb{R}^d$ of $I_t^i$ is computed by:

{
\small
\setlength\abovedisplayskip{-3pt}
\setlength\belowdisplayskip{5pt}
\begin{equation}
\begin{split}
    f(I_t^i)=\Theta_{l}I_t^i &+ \sum_{j\in \mathcal{N}_p(i)} W_p(i,j) \odot \Theta_{p}I_t^j \\
                             &+ \sum_{j\in \mathcal{N}_s(i)} W_s(i,j) \odot \Theta_{s}I_t^j \\
                             &+ \sum_{j\in \mathcal{N}_c(i)} W_c(i,j) \odot \Theta_{c}I_t^j,
\end{split}
\label{eq:gc}
\end{equation}
}%
where $\odot$ is Hadamard product and ${\Theta}=\{\Theta_l, \Theta_p, \Theta_s, \Theta_c\}$. Specifically, $\Theta_{l}I_t^i$ is the self-loop for all graphs and $\Theta_l$ is the learnable parameters. $\Theta_{p}$ denotes the parameters of the physical graph $\mathcal{G}_p$ and $\mathcal{N}_p(i)$ represents the neighbor set of node $i$ in $\mathcal{G}_p$. Other notations $\Theta_{s}$, $\Theta_{c}$, $\mathcal{N}_s(i)$ and $\mathcal{N}_c(i)$ have similar semantic meanings. $d$ is the dimensionality of feature $f(I_t^i)$. In this manner, a node can dynamically receive information from some highly-correlated neighbor nodes.
For convenience, the graph convolution in Eq.\ref{eq:gc} is abbreviated as $I_t * {\Theta}$ in the following.

\begin{figure}
    \centering
    \includegraphics[width=0.545\columnwidth]{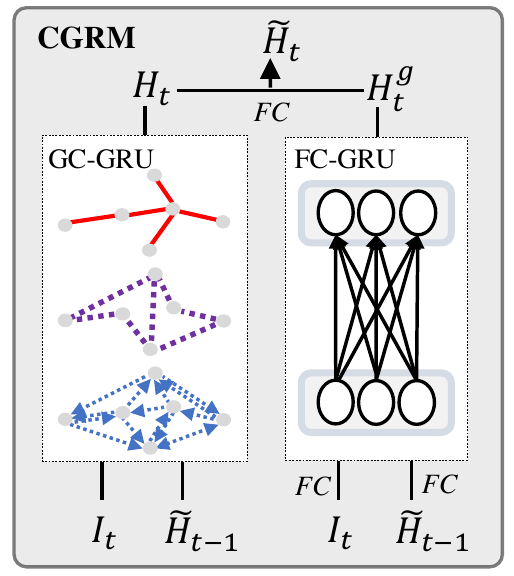}
    \vspace{-2mm}
    \caption{Illustration of Collaborative Gated Recurrent Module (CGRM) for local and global feature fusion. Specifically, our CGRUs consists of a Graph Convolution Gated Recurrent Unit (GC-GRU) and a Fully-Connected Gated Recurrent Unit (FC-GRU). FC denotes a fully-connected layer.}
    \label{fig:CGRM}
\end{figure}

Since the above-mentioned operation is conducted on spatial view, we embed the physical-virtual graph convolution in a Gated Recurrent Unit to learn spatial-temporal features. Specifically, the reset gate $R_t = \{R_t^1, R_t^2,...,R_t^N\}$, update gate $Z_t = \{Z_t^1,Z_t^2,...,Z_t^N\} $, new information $N_t = \{N_t^1,N_t^2,...,N_t^N\}$ and hidden state $H_t = \{H_t^1,H_t^2,...,H_t^N\}$ are computed by:
{\setlength\abovedisplayskip{6pt}
\setlength\belowdisplayskip{6pt}
\begin{equation}
\begin{split}
    R_t &= \sigma({\Theta}_{rx} * I_t + {\Theta}_{rh} * H_{t-1} + b_{r}) \\
    Z_t &= \sigma({\Theta}_{zx} * I_t + {\Theta}_{zh} * H_{t-1} + b_{z}) \\
    N_t &= \tanh\{{\Theta}_{nx} * I_t  + R_t \odot ({\Theta}_{nh} * H_{t-1} + b_{n})\} \\
    H_t &= (1 - Z_t) \odot N_t + Z_t \odot H_{t-1} \\
\end{split}
\label{eq:gc_gru}
\end{equation}}%
where $\sigma$ is the sigmoid function and $H_{t-1}$ is the hidden state at last iteration. ${\Theta}_{rx}$ denotes the graph convolution parameters between $H_t$ and $\mathbf{X}_t$, while $\Theta_{rh}$ denotes the parameters between $R_t$ and $H_{t-1}$. Other parameters ${\Theta}_{zx}$, ${\Theta}_{zh}$, ${\Theta}_{nx}$ and ${\Theta}_{nh}$ have similar meanings. $b_r$, $b_{z}$ and $b_{n}$ are bias terms. The feature dimensionality of $R_t^i$, $Z_t^i$, $N_t^i$ and $H_t^i$ are also set to $d$. For convenience, we denote the operation of Eq.\ref{eq:gc_gru} as:
{\begin{equation}
    H_t = \text{GC-GRU}(I_t, H_{t-1}) \\
\label{eq:gc_gru_short}
\end{equation}}%
Thanks to this GC-GRU, we can effectively learn spatial-temporal features from the ridership data of metro systems.

\subsection{Local-Global Feature Fusion}\label{sec:fusion}
In previous works~\cite{liu2019contextualized,fang2019gstnet}, global features have been proven to be also useful for traffic state prediction. However, the proposed GC-GRU conducts convolution on local space and fails to capture the global context. To address this issue, we apply a Fully-Connected Gated Recurrent Unit (FC-GRU) to learn the global evolution features of all stations and generate a comprehensive feature by fusing the output of GC-GRU and FC-GRU. The developed fusion module is termed as Collaborative Gated Recurrent Module (CGRM) in this work and its architecture is shown in Fig.\ref{fig:CGRM}.

Specifically, the inputs of CGRM are $I_t$ and $\tilde{H}_{t-1}$, where $\tilde{H}_{t-1}$ is the output of last iteration. For GC-GRU, rather than take the original $H_t$ as input, it utilizes the accumulated information in $\tilde{H}_{t-1}$ to update hidden state, thus Eq.\ref{eq:gc_gru_short} becomes:
{\begin{equation}
    H_t = \text{GC-GRU}(I_t, \tilde{H}_{t-1}) \\
\end{equation}}%
For FC-GRU, we first transform $I_t$ and $\tilde{H}_{t-1}$ to an embedded $I_t^g \in \mathbb{R}^d$ and $H_{t-1}^g \in \mathbb{R}^d$ with two fully-connected (FC) layers. Then we feed $I_t^e$ and $H_{t-1}^e$ into a common GRU~\cite{chung2014empirical} implemented with fully-connection to generate a global hidden state $\tilde{H}_t^g \in \mathbb{R}^d$, which can be expressed as:
{\begin{equation}
\begin{split}
    I_t^e &= \text{FC}(I_t), {~~} H_{t-1}^e = \text{FC}(\tilde{H}_{t-1}), \\
    H_t^g &= \text{FC-GRU}(I_t^e, H_{t-1}^e), \\
\end{split}
\end{equation}}%
Finally, we incorporate $H_t$ and $H_t^g$ to generate a comprehensive hidden state $\tilde{H}_t = \{\tilde{H}_t^1,\tilde{H}_t^2,...,\tilde{H}_t^N\}$ with a fully-connected
layer:
{\begin{equation}
    \tilde{H}_t^i = \text{FC}(H_t^i \oplus H_t^g),
\end{equation}}%
where $\oplus$ denotes an operation of feature concatenation. $\tilde{H}_t$ contains the local and global context of ridership, and we has proved its effectiveness in Section~\ref{sec:LG_effect}.

\subsection{Physical-Virtual Collaboration Graph Network}
In this section, we apply the above-mentioned CGRUs to construct a Physical-Virtual Collaboration Graph Network (PVCGN) for station-level metro ridership prediction. Following previous works~\cite{bai2019stg2seq,zhang2019multistep,pan2019urban,chen2020multi,zhang2020spatio,zheng2020gman}, we also adopt the Seq2Seq architecture to develop our framework, whose architecture is shown in Fig.\ref{fig:arch}.

Specifically, PVCGN consists of an encoder and a decoder, both of which contain two CGRMs. {\textbf{In encoder}}, the ridership data $\{\bm{X}_{t-n+1},\bm{X}_{t-n+2},...,\bm{X}_{t}\}$ are sequentially fed into CGRMs to accumulate the historical information. At iteration $i$, the bottom CGRM takes ${\bm{X}_{t-n+i}}$ as input and its output hidden state is fed into the above CGRM for high-level feature learning. In particular, the initial hidden states of both CGRMs at the first iteration are set to zero.
{\textbf{In decoder}}, at the first iteration, the input data is also set to zero and the final hidden states of encoder are used to initialize the hidden states of the decoder. The future ridership $\hat{\bm{X}}_{t+1}$ is predicted by feeding the output hidden state of the above CGRM into a fully-connected layer. At iteration $i{~~}(i\geq2)$, the bottom CGRM takes $\hat{\bm{X}}_{t+i-1}$ as input data and the above CGRM also applies a fully-connected layer to forecast $\hat{\bm{X}}_{t+i}$.
Finally, we can obtain a future ridership sequence $\{\hat{\bm{X}}_{t+1},\hat{\bm{X}}_{t+2},..., \hat{\bm{X}}_{t+m}\}$.

\begin{table}[t]
  \caption{The Overview of SHMetro and HZMetro datasets. ``\# Station" denotes the number of metro stations. ``M'' and ``min'' are the abbreviations of ``million'' and ``minute'', respectively.}
  \vspace{-2mm}
\newcommand{\tabincell}[2]{\begin{tabular}{@{}#1@{}}#2\end{tabular}}
  \centering
    \begin{tabular}{c|c|c}
    \hline
    \multicolumn{1}{c|}{\textbf{Dataset}}  & \textbf{SHMetro} & \textbf{HZMetro} \\
    \hline\hline
     City & Shanghai, China & Hangzhou, China \\
    \hline
     \# Station &  288 & 80 \\
     \hline
     \# Physical Edge & 958 & 248 \\
     \hline
     Ridership/Day & 8.82 M & 2.35 M \\
     \hline
     Time Interval & 15 min & 15 min \\
     \hline
    {~~Training Timespan} & 7/01/2016 - 8/31/2016 & 1/01/2019 - 1/18/2019 \\
    {Validation Timespan} & 9/01/2016 - 9/09/2016 & 1/19/2019 - 1/20/2019 \\
    {~~~~Testing Timespan} & 9/10/2016 - 9/30/2016 & 1/21/2019 - 1/25/2019 \\
    \hline
    \end{tabular}
  \label{tab:dataset_detail}
\end{table}

\section{Experiments}\label{sec:experiment}
In this section, we first introduce the settings of experiments (e.g., dataset construction, implementation details, and evaluation metrics). Then, we compare the proposed PVCGN with eight representative approaches under various scenarios. Finally, we conduct extensive internal analyses to verify the effectiveness of each component in our method.

\subsection{Experiments Settings}
\subsubsection{\textbf{Dataset Construction}}
Since there are few public benchmarks for metro ridership prediction, we collect a mass of trip transaction records from two real-world metro systems and construct two large-scale datasets, which are termed as HZMetro and SHMetro respectively. The overviews of these two datasets are summarized in Table~\ref{tab:dataset_detail}.

\textit{SHMetro}: This dataset was built based on the metro system of Shanghai, China. A total of 811.8 million transaction records were collected from Jul. 1st 2016 to Sept. 30th 2016, with 8.82 million ridership per day. Each record contains the information of passenger ID, entry/exit station and the corresponding timestamps. In this time period, 288 metro stations were operated normally and they were connected by 958 physical edges. For each station, we measured its inflow and outflow of every 15 minutes by counting the number of passengers entering or exiting the station. The ridership data of the first two months and that of the last three weeks are used for training and testing, while the ridership data of the remaining days are used for validation.

\textit{HZMetro}: This dataset was created with the transaction records of the Hangzhou metro system collected in January 2019. With 80 operational stations and 248 physical edges, this system has 2.35 million ridership per day. The time interval of this dataset is also set to 15 minutes. Similar to SHMetro, this dataset is divided into three parts, including a training set (Jan. 1st - Jan. 18th), a validation set (Jan. 19th - Jan. 20th) and a testing set (Jan. 21th - Jan. 25th).

\subsubsection{\textbf{Implementation Details}}\label{sec:implementation}
Since the physical graph has a well-defined topology, we only introduce the details of the two virtual graphs in this section.
In SHMetro dataset, to reduce the computational cost of GCN, for each station, we only select the top ten stations with high similarity scores or correlation rates to construct virtual graphs, thereby both the similarity graph and correlation graph have 2880 edges. In HSMetro dataset, as its station number is much smaller than that of SHMetro and the computational cost is not heavy, we can build more virtual edges to learn the complex patterns. Specifically, we determine the virtual edges by setting the similarity/correlation thresholds to 0.1/0.02, and the final similarity graph and correlation graph have 2502 and 1094 edges respectively.

We implement our PVCGN with the popular deep learning framework PyTorch~\cite{paszke2017automatic}. The lengths of input and output sequences are set to 4 simultaneously. The input data and the ground-truth of output are normalized with Z-score Normalization \footnote{\url{https://en.wikipedia.org/wiki/Standard_score}} before being fed into the network. The filter weights of all layers are initialized by Xavier~\cite{glorot2010understanding}. The batch size is set to 8 for SHMetro and 32 for HZmetro. The feature dimensionality $d$ is set to 256. The initial learning rate is 0.001 and its decay ratio is 0.1. We apply Adam~\cite{kingma2014adam} to optimize our PVCGN for 200 epochs by minimizing the mean absolute error between the predicted results and the corresponding ground-truths. On each benchmark, we train the proposed PVCGN with the training set and determine the model's hyper-parameters with the validation set. Finally, the well-trained model is evaluated on the testing set.

\subsubsection{\textbf{Evaluation Metrics}}
Following previous works~\cite{li2018diffusion,zhao2019t}, we evaluate the performance of methods with Root Mean Square Error (RMSE), Mean Absolute Error (MAE) and Mean Absolute Percentage Error (MAPE), which are defined as:
\begin{equation}
\begin{split}
    \text{RMSE} &= \sqrt{{\frac{1}{n}}{\sum_{i=1}^n{(\hat{X_i} - X_i)}^2}}, \\
    \text{MAE}  &= {\frac{1}{n}}{\sum_{i=1}^n |\hat{X_i} - X_i|}, \\
    \text{MAPE} &= {\frac{1}{n}}{\sum_{i=1}^n \frac{|\hat{X_i} - X_i|}{X_i}}
\end{split}
\end{equation}
where $n$ is the number of testing samples. ${\hat{X_i}}$ and ${X_i}$ denote the predicted ridership and the ground-truth ridership, respectively. Notice that ${\hat{X_i}}$ and ${X_i}$ have been transformed back to the original scale with an inverted Z-score Normalization. As mentioned in Section~\ref{sec:implementation}, our PVCGN is developed to predict the metro ridership of next four time intervals. In the following experiments, we would measure the errors of each time intervals separately.

\begin{table*}[t]
 \caption{Quantitative comparison on the whole SHMetro Dataset. Our PVCGN outperforms existing methods in all metrics.}
  \vspace{0mm}
\newcommand{\tabincell}[2]{\begin{tabular}{@{}#1@{}}#2\end{tabular}}
  \centering
  \resizebox{18.2cm}{!} {
    \begin{tabular}{c|c|c|c|c|c|c|c|c|c|c|c|c|c}
    \hline
    Time & Metric & {\textbf{HA}}	& {\textbf{RF}} & {\textbf{GBDT}} & {\textbf{MLP}} & {\textbf{LSTM}} & {\textbf{GRU}} & {\textbf{ASTGCN}} & {\textbf{STG2Seq}} & {\textbf{DCRNN}} & {\textbf{GCRNN}} & {\textbf{Graph-WaveNet}} & {\textbf{PVCGN}} \\
    \hline
    \hline
    \multirow{3}{*}{15 min}
     & \textit{RMSE}  & 136.97   & 66.63   & 62.59   & 48.71   & 55.53   & 52.04   & 66.49   & 47.19 & 46.02   & 46.09   & 46.98   & 44.97   \\
     & \textit{MAE~~} &  48.26   & 34.37   & 32.72   & 25.16   & 26.68   & 25.91   & 32.29   & 24.98 & 24.04   & 24.26   & 24.91   & 23.29   \\
     & \textit{MAPE}  &  31.55\% & 24.09\% & 23.40\% & 19.44\% & 18.76\% & 18.87\% & 21.90\% & 23.26\% & 17.82\% & 18.06\% & 20.05\% & 16.83\%\\
    \hline
    \multirow{3}{*}{30 min}
     & \textit{RMSE}  & 136.81  & 88.03   & 82.32   & 51.80   & 57.37   & 54.02   & 98.76   & 50.58 & 49.90   & 50.12   & 51.64   & 47.83   \\
     & \textit{MAE~~} & 47.88   & 41.37   & 39.50   & 26.15   & 27.25   & 26.39   & 39.28   & 26.17  & 25.23   & 25.42   & 26.53   & 24.16   \\
     & \textit{MAPE}  & 31.49\% & 28.89\% & 28.17\% & 20.38\% & 19.04\% & 19.20\% & 25.63\%	& 26.79\% & 18.35\% & 18.73\% & 20.38\% & 17.23\% \\
    \hline
    \multirow{3}{*}{45 min}
     & \textit{RMSE}  & 136.45  & 118.65  & 113.95  & 57.06   & 60.45   & 56.97   & 133.28  & 52.68 & 54.92   & 54.87   & 58.50   & 52.02  \\
     & \textit{MAE~~} & 47.26   & 50.91   & 49.14   & 27.91   & 28.08   & 27.17   & 46.59   & 26.75 & 26.76   & 26.92   & 28.78   & 25.33   \\
     & \textit{MAPE}  & 31.27\% & 41.34\% & 40.76\% & 22.20\% & 19.61\% & 19.84\% & 29.45\% & 28.49\% & 19.30\% & 19.81\% & 21.99\% & 17.92\% \\
    \hline
    \multirow{3}{*}{60 min}
     & \textit{RMSE}  & 135.72  & 143.5   & 137.5   & 63.33   & 63.41   & 59.91   & 154.95  & 56.81 & 58.83   & 58.67   & 65.08   & 55.27   \\
     & \textit{MAE~~} & 46.40   & 59.15   & 57.31   & 29.92   & 28.94   & 28.08   & 51.33   & 28.22 & 28.01   & 28.18   & 30.90   & 26.29   \\
     & \textit{MAPE}  & 30.80\% & 52.91\% & 52.60\% & 23.96\% & 20.59\% & 21.03\% & 32.35\% & 34.30\% & 20.44\% & 21.07\% & 24.36\% & 18.69\% \\
    \hline
    \end{tabular}
  }
  \label{tab:whole_SHMetro}
\end{table*}

\begin{table*}[t]
 \caption{Quantitative comparison on the whole HZMetro Dataset. Our PVCGN outperforms existing methods in all metrics.}
  \vspace{0mm}
\newcommand{\tabincell}[2]{\begin{tabular}{@{}#1@{}}#2\end{tabular}}
  \centering
  \resizebox{18.2cm}{!} {
    \begin{tabular}{c|c|c|c|c|c|c|c|c|c|c|c|c|c}
    \hline
    Time & Metric & {\textbf{HA}}	& {\textbf{RF}} & {\textbf{GBDT}} & {\textbf{MLP}} & {\textbf{LSTM}} & {\textbf{GRU}} & {\textbf{ASTGCN}} & {\textbf{STG2Seq}} & {\textbf{DCRNN}} & {\textbf{GCRNN}} & {\textbf{Graph-WaveNet}} & {\textbf{PVCGN}} \\
    \hline
    \hline
    \multirow{3}{*}{15 min}
     & \textit{RMSE}  & 64.19   & 53.52   & 51.50   & 46.55   & 45.30   & 45.10   & 46.19   & 39.52 & 40.39   & 40.24   & 40.78   & 37.76   \\
     & \textit{MAE~~} & 36.37   & 32.19   & 30.88   & 26.57   & 25.76   & 25.69   & 27.34   & 23.8 & 23.76   & 23.84   & 24.07   & 22.68   \\
     & \textit{MAPE}  & 19.14\% & 18.34\% & 17.60\% & 16.26\% & 14.91\% & 15.13\% & 15.05\%	& 17.09\% & 14.00\% & 14.08\% & 14.27\% & 13.70\% \\
    \hline
    \multirow{3}{*}{30 min}
     & \textit{RMSE}  & 64.10   & 64.54   & 61.94   & 47.96   & 45.52   & 45.26   & 46.16   & 40.72 & 42.57   & 41.95   & 42.80   & 39.34   \\
     & \textit{MAE~~} & 36.37   & 38.00   & 36.48   & 27.44   & 26.01   & 25.93   & 27.74   & 24.72 & 25.22   & 25.14   & 25.48   & 23.33   \\
     & \textit{MAPE}  & 19.31\% & 21.46\% & 20.49\% & 17.10\% & 15.10\% & 15.35\% & 15.56\% & 19.51\% & 14.99\% & 14.86\% & 15.23\% & 13.81\% \\
    \hline
    \multirow{3}{*}{45 min}
     & \textit{RMSE}  & 63.92   & 80.06   & 76.70   & 50.66   & 46.30   & 46.13   & 46.79   & 43.36 & 46.26   & 45.53   & 45.84   & 40.95   \\
     & \textit{MAE~~} & 36.23   & 45.78   & 44.12   & 28.79   & 26.38   & 26.36   & 28.20   & 25.98 & 26.97   & 26.82   & 27.15   & 24.22   \\
     & \textit{MAPE}  & 19.57\% & 26.51\% & 25.75\% & 19.01\% & 15.40\% & 15.79\% & 16.48\% & 23.59\% & 16.19\% & 16.05\% & 17.34\% & 14.45\% \\
    \hline
    \multirow{3}{*}{60 min}
     & \textit{RMSE}  & 63.72   & 94.29   & 91.21   & 54.62   & 47.53   & 47.69   & 49.70   & 46.05 & 49.35   & 50.28   & 49.89   & 42.61   \\
     & \textit{MAE~~} & 35.99   & 52.95   & 51.10   & 30.52   & 26.76   & 26.98   & 28.85   & 26.5 & 28.47   & 28.75   & 29.14   & 24.93   \\
     & \textit{MAPE}  & 20.01\% & 37.12\% & 38.10\% & 22.56\% & 16.34\% & 17.20\% & 17.75\% & 27.93\% & 18.16\% & 17.89\% & 19.37\% & 15.49\% \\
    \hline
    \end{tabular}
  }
  \label{tab:whole_HZMetro}
\end{table*}

\subsection{Comparison with State-of-the-Art Methods}\label{sec:SOTA_Comp}
In this section, we compare our PVCGN with nine basic and advanced methods under various scenarios (e.g., the comparison on the whole testing sets, comparison on rush hours and comparison on high-ridership stations).
These methods can be classified into three categories, including: {\bf{(i)}} three traditional time series models, {\bf{(ii)}} three general deep learning models and {\bf{(iii)}} five recently-proposed graph networks. The details of these methods are described as following:
\begin{itemize}
  \item \textbf{Historical Average (HA):} HA is a seasonal-based baseline that forecasts the future ridership by averaging the riderships of the corresponding historical periods. For instance, the ridership at interval 9:00-9:15 on a specific Monday is predicted as the average ridership of the corresponding time intervals of the previous $k$ Mondays. The variate $k$ is set to 4 on SHMetro and 2 on SHMetro.
  \item \textbf{Random Forest (RF):} RF is a machine learning technique for both regression and classification problems that operates by constructing a multitude of decision trees. Sklearn is used to implement this method. The number of trees is set to 10 and the maximum depth is automatically expanded until all leaves are pure or until all leaves contain less than 2 samples.
  \item \textbf{Gradient Boosting Decision Trees (GBDT):} GBDT is a weighted ensemble method that consists of a series of weak estimators. We implement this method with Sklearn. The number of boosting stages is set to 100 and the maximum depth of each estimator is 4. Gradient descent optimizer is applied to minimize the loss function.
  \item \textbf{Multiple Layer Perceptron (MLP):} This model consists of two fully-connected layers with 256 and $2{\times}4{\times}s$ neurons respectively, where $s$ is the number of stations. It takes as input the riderships of all stations of previous $n$ time intervals and predicts the ridership of all stations of the next $m$ time intervals simultaneously. Its hyper-parameters are the same as ours.
  \item \textbf{Long Short-Term Memory (LSTM):} This network is a simple Seq2Seq model and its core module consists of two fully-connected LSTM layers. The hidden size of each LSTM layer is set to 256. Its hyper-parameters are the same as ours.
  \item \textbf{Gated Recurrent Unit (GRU):} With the similar architecture of the previous model, this network replaces the original LSTM layers with GRU layers. The hidden size of GRU is also set to 256. Its hyper-parameters are the same as ours.

  \begin{table*}[t]
 \caption{Quantitative comparison during rush hours on SHMetro Dataset. The rush time refers to 7:30-9:30 and 17:30-19:30.}
  \vspace{0mm}
\newcommand{\tabincell}[2]{\begin{tabular}{@{}#1@{}}#2\end{tabular}}
  \centering
  \resizebox{18.2cm}{!} {
    \begin{tabular}{c|c|c|c|c|c|c|c|c|c|c|c|c|c}
    \hline
    Time & Metric & {\textbf{HA}}	& {\textbf{RF}} & {\textbf{GBDT}} & {\textbf{MLP}} & {\textbf{LSTM}} & {\textbf{GRU}} & {\textbf{ASTGCN}} & {\textbf{STG2Seq}} & {\textbf{DCRNN}} & {\textbf{GCRNN}} & {\textbf{Graph-WaveNet}} & {\textbf{PVCGN}} \\
    \hline
    \hline
    \multirow{3}{*}{15 min}
     & \textit{RMSE}  &  255.63  & 108.09  & 100.49  & 64.95   & 75.78   & 69.92   & 91.98   & 66.29   & 67.50   & 66.21   & 68.41   & 65.04   \\
     & \textit{MAE~~} &  96.23   & 56.20   & 53.36   & 36.42   & 39.49   & 37.27   & 47.94   & 38.05   & 37.92   & 37.94   & 39.17   & 36.46   \\
     & \textit{MAPE}  &  46.74\% & 20.06\% & 19.20\% & 14.47\% & 13.87\% & 13.58\% & 21.45\% & 14.90\% & 13.93\% & 14.07\% & 14.14\% & 13.16\% \\
    \hline
    \multirow{3}{*}{30 min}
     & \textit{RMSE}  & 270.74  & 161.33  & 149.34  & 69.97   & 77.24   & 72.19   & 153.92  & 72.45   & 73.07   & 73.63   & 78.98   & 68.85   \\
     & \textit{MAE~~} & 99.18   & 75.06   & 72.13   & 38.24   & 39.97   & 37.73   & 62.41   & 40.13   & 40.16   & 40.26   & 43.54   & 37.77   \\
     & \textit{MAPE}  & 47.10\% & 23.72\% & 22.85\% & 14.65\% & 14.02\% & 13.61\% & 28.03\% & 15.49\% & 14.33\% & 14.44\% & 14.79\% & 13.41\% \\
    \hline
    \multirow{3}{*}{45 min}
     & \textit{RMSE}  & 265.61  & 231.29  & 222.53  & 74.30   & 79.05   & 72.70   & 204.12  & 73.04   & 79.42   & 79.88   & 87.66   & 73.85   \\
     & \textit{MAE~~} & 95.56   & 98.57   & 97.18   & 39.58   & 39.77   & 37.33   & 73.06   & 39.79   & 41.92   & 41.65   & 46.26   & 38.84   \\
     & \textit{MAPE}  & 45.39\% & 29.61\% & 28.97\% & 15.42\% & 14.45\% & 14.04\% & 32.92\% & 15.64\% & 15.29\% & 15.29\% & 15.95\% & 14.06\% \\
    \hline
    \multirow{3}{*}{60 min}
     & \textit{RMSE}  & 248.93  & 284.02  & 271.83  & 75.72   & 77.13   & 69.71   & 220.25  & 75.46   & 79.98   & 81.37   & 90.19   & 74.41   \\
     & \textit{MAE~~} & 87.10   & 115.13  & 113.73  & 39.53   & 38.23   & 35.68   & 74.38   & 39.67   & 41.23   & 41.27   & 46.35   & 38.12   \\
     & \textit{MAPE}  & 42.48\% & 36.00\% & 35.54\% & 16.59\% & 15.27\% & 14.81\% & 34.81\% & 17.09\% & 16.58\% & 16.62\% & 17.66\% & 15.08\% \\
    \hline
    \end{tabular}
  }
  \label{tab:SHMetro_rush}
\end{table*}

\begin{table*}[t]
 \caption{Quantitative comparison during rush hours on HZMetro Dataset. The rush time refers to 7:30-9:30 and 17:30-19:30.}
  \vspace{0mm}
\newcommand{\tabincell}[2]{\begin{tabular}{@{}#1@{}}#2\end{tabular}}
  \centering
  \resizebox{18.2cm}{!} {
    \begin{tabular}{c|c|c|c|c|c|c|c|c|c|c|c|c|c}
    \hline
    Time & Metric & {\textbf{HA}}	& {\textbf{RF}} & {\textbf{GBDT}} & {\textbf{MLP}} & {\textbf{LSTM}} & {\textbf{GRU}} & {\textbf{ASTGCN}} & {\textbf{STG2Seq}} & {\textbf{DCRNN}} & {\textbf{GCRNN}} & {\textbf{Graph-WaveNet}} & {\textbf{PVCGN}} \\
    \hline
    \hline
    \multirow{3}{*}{15 min}
     & \textit{RMSE}  & 65.53   & 84.33   & 82.25   & 57.39   & 57.10  & 56.31   & 60.72   & 53.28   & 54.17   & 55.51   & 56.98    & 49.79  \\
     & \textit{MAE~~} & 40.63   & 52.07   & 51.60   & 35.77   & 35.27  & 35.23   & 36.82   & 35.03   & 35.08   & 35.68   & 37.19    & 32.63  \\
     & \textit{MAPE}  & 11.51\% & 15.24\% & 15.02\% & 10.96\% & 9.99\% & 10.12\% & 11.77\% & 10.73\% & 10.37\% & 10.36\% & 10.84\%  & 9.72\% \\
    \hline
    \multirow{3}{*}{30 min}
     & \textit{RMSE}  & 67.89   & 108.25  & 103.38  & 62.25   & 59.03   & 58.81   & 58.30   & 56.26   & 58.27   & 57.34   & 59.71    & 51.63  \\
     & \textit{MAE~~} & 42.08   & 65.97   & 63.94   & 37.58   & 36.45   & 36.59   & 35.48   & 36.96   & 37.48   & 37.31   & 38.94    & 33.30  \\
     & \textit{MAPE}  & 11.58\% & 17.56\% & 16.95\% & 10.80\% & 10.07\% & 10.10\% & 12.15\% & 10.95\% & 10.69\% & 10.54\% & 11.04\%  & 9.52\% \\
    \hline
    \multirow{3}{*}{45 min}
     & \textit{RMSE}  & 67.33   & 123.34  & 126.36  & 61.85   & 58.48   & 58.13   & 57.23   & 58.17   & 61.83   & 59.54   & 59.83    & 51.45  \\
     & \textit{MAE~~} & 41.63   & 74.91   & 75.46   & 37.09   & 35.72   & 35.59   & 34.59   & 37.10   & 37.95   & 37.58   & 38.75    & 32.73  \\
     & \textit{MAPE}  & 12.22\% & 19.58\% & 19.36\% & 11.30\% & 10.55\% & 10.36\% & 12.84\% & 11.72\% & 11.16\% & 11.16\% & 11.83\%  & 9.88\% \\
    \hline
    \multirow{3}{*}{60 min}
     & \textit{RMSE}  & 67.22   & 136.08  & 132.87  & 61.81   & 57.35   & 57.14   & 59.23   & 57.69   & 59.52   & 58.88   & 59.96    & 51.09   \\
     & \textit{MAE~~} & 40.72   & 75.40   & 74.39   & 36.13   & 34.19   & 34.01   & 33.59   & 35.64   & 36.27   & 35.94   & 37.49    & 31.43   \\
     & \textit{MAPE}  & 13.21\% & 20.81\% & 20.54\% & 12.16\% & 11.23\% & 11.08\% & 13.68\% & 12.25\% & 11.94\% & 11.93\% & 12.35\%  & 10.43\% \\
    \hline
    \end{tabular}
  }
  \label{tab:HZMetro_rush}
\end{table*}

  \item \textbf{Attention Based Spatial-Temporal Graph Convolutional Networks (ASTGCN\cite{guo2019attention}):} In this network, a spatial-temporal attention mechanism and a spatial-temporal convolution are developed to simultaneously capture the spatial patterns and temporal patterns from traffic data. Based on its official code, we apply this model to metro ridership prediction.
  \item \textbf{Spatial-temporal Graph to Sequence Model (STG2Seq \cite{bai2019stg2seq}):} This method applies a hierarchical graph convolutional structure to capture both spatial and temporal correlations simultaneously. It consists of a short-term encoder, a long-term encoder, and an attention-based fusion module. Based on the official code, this method is re-implemented for metro ridership prediction.
  \item \textbf{Diffusion Convolutional Recurrent Neural Network (DCRNN \cite{li2018diffusion}):} As a deep learning framework specially designed for traffic forecasting, DCRNN captures the spatial dependencies using bidirectional random walks on graphs and learns the temporal dependencies with an encoder-decoder architecture. We implement this method based on its official code.
  \item \textbf{Graph Convolutional Recurrent Neural Network (GCRNN):} The architecture and setting of this method are very similar to these of DCRNN. The main difference is that GCRNN replaces diffusion convolutional layers with $K$=3 order ChebNets~\cite{kipf2016semi} based on spectral graph convolutions.
  \item \textbf{Graph-WaveNet \cite{wu2019graph}:} This method develops an adaptive dependency matrix to capture the hidden spatial dependency and utilizes a stacked dilated 1D convolution component to handle very long sequences. We implement this method with its official code.
\end{itemize}

\subsubsection{\textbf{Comparison on the Whole Testing Sets}}
We first compare the performance of all comparative methods on the whole testing sets (including all time intervals and all metro stations). Their performance on SHMetro and HZMetro datasets are summarized in Table~\ref{tab:whole_SHMetro} and Table~\ref{tab:whole_HZMetro}, respectively.
We can see that the baseline HA obtains unacceptable MAPE at all time intervals (about 31\% on SHMetro and 20\% on HZMetro). Compared with HA, RF and GBDT can get better results at the first time interval. However, with the increment of time, their MAPEs gradually become worse and even larger than that of HA, since these two traditional models have weak abilities to learn the ridership distribution. By automatically learning deep features from data, those general neural networks (e.g., MLP, LSTM and GRU) can greatly improve the performance. For example, LSTM obtains a MAPE 18.76\% on SHMetro and 14.91\% on HZMetro when predicting the ridership at the first time interval, while GRU obtains a MAPE 21.03\% on SHMetro and 17.20\% on HZMetro for the prediction of the fourth time interval. Thanks to the advanced graph learning, DCRNN and GCRNN achieve competitive performance by reducing the MAPE to 17.82\% on SHMetro and to 14.00\% on HZMetro. However, these methods directly construct graphs based on physical topologies. To fully capture the ridership complex patterns, our PVCGN constructs physical/virtual graphs with the information of physical topologies and human domain knowledge, thereby achieving state-of-the-art performance. For example, our PVCGN improves at least 1\% in MAPE at different time intervals on SHMetro dataset. On HZmetro, PVCGN outperforms the existing best-performing models DCRNN, GCRNN and Graph-WaveNet with a large margin in all metrics. This comparison well demonstrates the superiority of the proposed PVCGN.

\subsubsection{\textbf{Comparison on Rush Hours}}
In this section, we focus on the ridership prediction of rush hours, since the accurate prediction results are very crucial for the metro scheduling during such time. In this work, the rush time is defined as 7:30-9:30 and 17:30-19:30. The performance of all methods are summarized in Table~\ref{tab:SHMetro_rush} and Table~\ref{tab:HZMetro_rush}.
We can observe that our PVCGN outperforms all comparative methods consistently on both datasets. On SHMetro, our PVCGN obtains a MAPE 13.16\% for the ridership prediction at the first time interval, while the MAPE of DCRNN and GCRNN are 13.93\% and 14.07\%, respectively. Other deep learning methods (such as MLP, LSTM and GRU) are relatively worse. For forecasting the ridership at the fourth time interval, our PVCGN achieves a MAPE 15.08\%, outperforming DCRNN and GCRNN with a relative improvement of at least 9.04\%.

\begin{table*}[t]
 \caption{Quantitative comparison of the top 1/4 high-ridership stations on SHMetro Dataset. We rerank all metro stations on the basis of their riderships and choose the top 1/4 stations for comparison.}
  \vspace{0mm}
\newcommand{\tabincell}[2]{\begin{tabular}{@{}#1@{}}#2\end{tabular}}
  \centering
  \resizebox{18.2cm}{!} {
    \begin{tabular}{c|c|c|c|c|c|c|c|c|c|c|c|c|c}
    \hline
    Time & Metric & {\textbf{HA}}	& {\textbf{RF}} & {\textbf{GBDT}} & {\textbf{MLP}} & {\textbf{LSTM}} & {\textbf{GRU}} & {\textbf{ASTGCN}} & {\textbf{STG2Seq}} & {\textbf{DCRNN}} & {\textbf{GCRNN}} & {\textbf{Graph-WaveNet}} & {\textbf{PVCGN}} \\
    \hline
    \hline
    \multirow{3}{*}{15 min}
     & \textit{RMSE}  & 242.87  & 111.31  & 103.94  & 80.72   & 94.74   & 87.40   & 114.77  & 86.19 & 84.04   & 86.09   & 76.93   & 74.80   \\
     & \textit{MAE~~} & 96.38   & 60.65   & 57.47   & 45.31   & 49.29   & 47.09   & 62.96   & 47.06 & 44.98   & 45.89   & 43.31   & 41.38   \\
     & \textit{MAPE}  & 27.82\% & 15.24\% & 14.80\% & 12.23\% & 12.39\% & 12.23\% & 17.20\% & 15.92\% & 13.76\% & 14.12\% & 11.68\% & 10.62\% \\
    \hline
    \multirow{3}{*}{30 min}
     & \textit{RMSE}  & 242.68  & 152.10  & 141.45  & 86.46   & 98.02   & 91.25   & 174.70  & 93.58 & 88.52   & 89.89   & 84.23   & 79.43   \\
     & \textit{MAE~~} & 95.83   & 75.66   & 72.31   & 47.58   & 50.52   & 48.27   & 78.83   & 49.6 & 46.80   & 47.50   & 46.32   & 43.05   \\
     & \textit{MAPE}  & 28.08\% & 20.25\% & 20.08\% & 13.62\% & 13.19\% & 13.32\% & 21.84\% & 18.06\% & 14.47\% & 14.82\% & 13.12\% & 11.46\% \\
    \hline
    \multirow{3}{*}{45 min}
     & \textit{RMSE}  & 242.22  & 208.91  & 201.11  & 96.13   & 103.88  & 96.89   & 236.82  & 95.28 & 97.75   & 96.81   & 96.35   & 87.32 \\
     & \textit{MAE~~} & 94.84   & 95.56   & 93.02   & 51.63   & 52.54   & 50.19   & 95.53   & 50.8 & 49.84   & 50.08   & 50.76   & 45.67 \\
     & \textit{MAPE}  & 28.11\% & 34.72\% & 35.40\% & 16.37\% & 14.12\% & 14.54\% & 27.08\% & 21.98\% & 16.32\% & 16.70\% & 15.50\% & 12.48\% \\
    \hline
    \multirow{3}{*}{60 min}
     & \textit{RMSE}  & 241.27  & 255.64  & 245.41  & 107.53  & 109.64  & 102.81  & 275.41  & 108.63 & 106.03  & 102.93  & 109.26  & 93.59 \\
     & \textit{MAE~~} & 93.41   & 112.77  & 110.44  & 56.42   & 54.64   & 52.50   & 106.60  & 55.46 & 53.30   & 52.79   & 55.74   & 48.02 \\
     & \textit{MAPE}  & 27.80\% & 53.09\% & 54.37\% & 19.48\% & 15.70\% & 16.41\% & 31.06\% & 27.99\% & 17.89\% & 18.16\% & 18.27\% & 13.61\% \\
    \hline
    \end{tabular}
  \label{tab:SHMetro_high}
  }
\end{table*}

\begin{table*}[t]
 \caption{Quantitative comparison of the top 1/4 high-ridership stations on HZMetro Dataset. We rerank all metro stations on the basis of their riderships and choose the top 1/4 stations for comparison.}
  \vspace{0mm}
\newcommand{\tabincell}[2]{\begin{tabular}{@{}#1@{}}#2\end{tabular}}
  \centering
  \resizebox{18.2cm}{!} {
    \begin{tabular}{c|c|c|c|c|c|c|c|c|c|c|c|c|c}
    \hline
    Time & Metric & {\textbf{HA}}	& {\textbf{RF}} & {\textbf{GBDT}} & {\textbf{MLP}} & {\textbf{LSTM}} & {\textbf{GRU}} & \textbf{ASTGCN} & {\textbf{STG2Seq}} & {\textbf{DCRNN}} & {\textbf{GCRNN}} & {\textbf{Graph-WaveNet}} & {\textbf{PVCGN}} \\
    \hline
    \hline
    \multirow{3}{*}{15 min}
     & \textit{RMSE}  & 111.26  & 82.98   & 79.23   & 77.07   & 75.19   & 74.57   & 75.10   & 65.41 & 66.18   & 65.29   & 65.87    & 60.56\\
     & \textit{MAE~~} & 70.30   & 54.73   & 52.28   & 45.95   & 45.28   & 44.81   & 48.78   & 40.97 & 41.22   & 40.93   & 40.72    & 38.29 \\
     & \textit{MAPE}  & 16.36\% & 14.47\% & 13.80\% & 11.68\% & 11.49\% & 11.45\% & 12.42\% & 10.88\% & 10.72\% & 10.59\% & 10.39\%  & 9.97\%\\
    \hline
    \multirow{3}{*}{30 min}
     & \textit{RMSE}  & 111.01  & 99.84   & 95.27   & 79.30   & 75.48   & 74.75   & 74.99   & 66.94 & 69.36   & 67.29   & 68.92    & 63.77\\
     & \textit{MAE~~} & 70.19   & 64.59   & 62.02   & 47.80   & 45.86   & 45.24   & 49.32   & 43.02 & 43.92   & 43.07   & 43.21    & 39.93 \\
     & \textit{MAPE}  & 16.52\% & 17.49\% & 16.80\% & 12.31\% & 11.73\% & 11.80\% & 12.80\% & 11.90\% & 11.51\% & 11.42\% & 11.40\%  & 10.34\%\\
    \hline
    \multirow{3}{*}{45 min}
     & \textit{RMSE}  & 110.64  & 125.47  & 119.12  & 82.86   & 76.80   & 76.12   & 75.49   & 71.95 & 75.16   & 72.42   & 72.96    & 65.99\\
     & \textit{MAE~~} & 69.86   & 79.09   & 75.98   & 50.36   & 46.62   & 46.17   & 49.76   & 45.68 & 46.99   & 45.93   & 45.71    & 41.75 \\
     & \textit{MAPE}  & 16.93\% & 22.69\% & 21.97\% & 14.43\% & 12.09\% & 12.42\% & 13.59\% & 13.90\% & 12.74\% & 12.46\% & 13.32\%  & 11.28\% \\
    \hline
    \multirow{3}{*}{60 min}
     & \textit{RMSE}  & 110.34  & 148.45  & 145.35  & 89.47   & 79.27   & 79.11   & 79.80   & 78.33 & 79.79   & 80.34   & 79.70   & 69.25 \\
     & \textit{MAE~~} & 69.44   & 92.23   & 89.36   & 53.96   & 47.48   & 47.60   & 50.55   & 48.11 & 49.50   & 49.40   & 49.77   & 43.16 \\
     & \textit{MAPE}  & 17.64\% & 33.21\% & 32.26\% & 20.03\% & 13.41\% & 14.25\% & 14.99\% & 16.97\% & 15.00\% & 14.74\% & 16.56\% & 12.54\% \\
    \hline
    \end{tabular}
  }
  \label{tab:HZMetro_high}
\end{table*}

There exists a similar situation of performance comparison on HZMetro. For example, obtaining a MAPE 9.72\% for the ridership at the first time interval, our PVCGN is undoubtedly better than DCRNN and GCRNN, the MAPE of which are 10.37\% and 10.36\%, respectively. When predicting the ridership at the fourth time interval, our PVCGN achieves a very impressive MAPE 10.43\%, while DCRNN and GCRNN suffer from serious performance degradation. For instance, their MAPE rapidly increase to 11.94\% and to 11.93\%, respectively. In summary, the extensive experiments on SHMetro and HZMetro dataset show the effectiveness and robustness of our method during rush hours.

\subsubsection{\textbf{Comparison on High-Ridership Stations}}
Except for the prediction during rush hours, we also pay attention to the prediction of some stations with high ridership, since the demand of such stations should be prioritized in real-world metro systems. In this section, we first rerank all metro stations based on their historical ridership of training set and conduct choose comparison on the top 1/4 high-ridership stations.
The performance on SHMetro is summarized in Table~\ref{tab:SHMetro_high} and we can observe that our PVGCN ranks the first place in performance among all comparative methods. When forecasting the ridership during the next 15 minutes, PVGCN achieves an RMSE 74.80 and a MAPE 10.62\%. By contrast, the best RMSE and MAPE of other methods are 80.72 and 12.23\%. As the prediction time increases to 60 minutes, our PVGCN can still obtain the best result (e.g., 13.61\% in MAPE), while the MAPE of GCRNN significantly increases to 18.16\%.

As shown in Table~\ref{tab:HZMetro_high}, our PVGCN also achieves impressive performance on HZMetro dataset. For the ridership prediction at the first time interval, the RMSE and MAPE of our PVCGN are 60.56 and 9.97\%, while the RMSE and MAPE of the existing best-performing method GCRNN are 65.29 and 10.59\%. When forecasting the ridership at the fourth time interval, our PVCGN has minor performance degradation. For example, its RMSE and MAPE respectively increase to 69.25 and 12.54\%. In the same situation, the RMSE and MAPE of GCRNN increase to 80.34 and 14.74\%. Therefore, we can conclude that our PVCGN is not only effective but also robust for the prediction on high-ridership stations.

\begin{table}[t]
    \caption{Running time (seconds) of different methods. All methods can achieve practical efficiencies.}
  \vspace{0mm}
\newcommand{\tabincell}[2]{\begin{tabular}{@{}#1@{}}#2\end{tabular}}
  \centering
    \begin{tabular}{c|c|c}
    \hline
    \tabincell{c}{Model} & \tabincell{c}{SHMetro} & \tabincell{c}{HZMetro}  \\
    \hline\hline
    LSTM & 0.00057 & 0.00050  \\
    \hline
    GRU & 0.00047 & 0.00050  \\
    \hline
    DCRNN & 0.0121 & 0.0156 \\
    \hline
    GCRNN & 0.0102 & 0.0126  \\
    \hline
    PVCGN & 0.2298 & 0.0503  \\
    \hline
    \end{tabular}
  \label{tab:time}
\end{table}

\begin{table*}[t]
  \caption{Performance of different variants of our PVCGN. The physical graph, similarity graph and correlation graph is abbreviated as ``P'', ``S'' and ``C'' respectively.}
  \vspace{0mm}
\newcommand{\tabincell}[2]{\begin{tabular}{@{}#1@{}}#2\end{tabular}}
  \centering
    \begin{tabular}{c|c|c|c|c|c|c||c|c|c|c|c}
    \hline
    \multirow{2}{*}{Time} & \multirow{2}{*}{Metric} & \multicolumn{5}{|c||}{\textbf{SHMetro}} & \multicolumn{5}{c}{\textbf{HZMetro}} \\
    \cline{3-12}
     & & P & P+S & P+C & S+C & P+S+C & P & P+S & P+C & S+C & P+S+C \\
     \hline
     \hline
    \multirow{3}{*}{15 min}
     & \textit{RMSE}  & 50.45   & 47.38   & 46.18   & 46.52   & 44.97   & 41.80   & 38.89    & 39.46   & 39.92   & 37.73   \\
     & \textit{MAE~~} & 25.89   & 24.16   & 23.88   & 23.74   & 23.29   & 24.81   & 23.23   & 23.34   & 23.84   & 22.69   \\
     & \textit{MAPE}  & 19.04\% & 17.13\% & 17.12\% & 16.94\% & 16.83\% & 14.84\% & 13.93\% & 14.08\% & 14.38\% & 13.72\% \\
    \hline
    \multirow{3}{*}{30 min}
     & \textit{RMSE}  & 58.09   & 50.86   & 50.29   & 50.18   & 47.83   & 45.31   & 40.63   & 41.26   & 41.59   & 39.38   \\
     & \textit{MAE~~} & 28.13   & 25.28   & 25.13   & 24.74   & 24.16   & 26.63   & 24.22   & 24.22   & 24.59   & 23.35   \\
     & \textit{MAPE}  & 20.19\% & 17.72\% & 17.73\% & 17.32\% & 17.23\% & 15.50\% & 14.49\% & 14.36\% & 14.60\% & 13.83\% \\
    \hline
    \multirow{3}{*}{45 min}
     & \textit{RMSE}  & 65.81   & 55.98   & 55.54   & 54.45   & 52.02   & 50.26   & 42.63   & 43.96   & 44.81   & 40.88   \\
     & \textit{MAE~~} & 30.51   & 26.90   & 26.68   & 26.01   & 25.33   & 29.02   & 25.31   & 25.42   & 25.91   & 24.23   \\
     & \textit{MAPE}  & 21.65\% & 18.66\% & 18.44\% & 18.03\% & 17.92\% & 16.76\% & 15.35\% & 15.26\% & 15.23\% & 14.48\% \\
    \hline
    \multirow{3}{*}{60 min}
     & \textit{RMSE}  & 73.06   & 60.08   & 60.59   & 58.93   & 55.27   & 56.32   & 44.46   & 44.93   & 45.49   & 42.51  \\
     & \textit{MAE~~} & 32.55   & 27.92   & 27.94   & 27.14   & 26.29   & 31.41   & 26.16   & 26.13   & 26.54   & 24.90   \\
     & \textit{MAPE}  & 23.43\% & 19.56\% & 19.30\% & 18.87\% & 18.69\% & 18.33\% & 16.31\% & 16.32\% & 16.69\% & 15.48\% \\
    \hline
    \end{tabular}
  \label{tab:different_graphs}
  \vspace{0mm}
\end{table*}

\subsubsection{\bf{Efficiency Comparison}}
Finally, we compare the inference efficiencies of five deep learning methods. Note that all methods are run on the same NVIDIA Titan-X GPU and their running time are summarized in Table~\ref{tab:time}. It can be seen that LSTM and GRU are the most efficient models, while GCRNN and GCRNN cost 0.0121$\sim$0.0156 seconds for each inference. With three graphs, our PVCGN can still achieve practical efficiencies. Specifically, PVCGN only requires 0.2298 seconds on SHMetro and 0.0503 seconds on HZMetro to forecast the citywide metro ridership in the next hour. In summary, all methods can run in real time and the inference efficiency is not the bottleneck of this task.

\subsection{Component Analysis}
\subsubsection{\textbf{Effectiveness of Different Graphs}}
The distinctive characteristic of our work is that we incorporate a physical graph and two virtual graphs into Gated Recurrent Units (GRU) to collaboratively capture the complex flow patterns. To verify the effectiveness of each graph, we implement five variants of PVCGN, which are described as follows:
\begin{itemize}
\item \textbf{P-Net:} This variant only utilizes the physical graph to implement the ridership prediction network;
\item \textbf{P+S-Net:} This variant is developed with the physical graph and the virtual similarity graph;
\item \textbf{P+C-Net:} Similar with P+S GRU, this variant is built with the physical graph and the correlation graph;
\item \textbf{S+C-Net:} Different with above variants that contain the physical graph, this variant is constructed only with the virtual similarity/correlation graphs;
\item \textbf{P+S+C-Net:} This network is the full model of the proposed PVCGN. It contains the physical graph and the two virtual graphs simultaneously.
\end{itemize}

The performance of all variants are summarized in Table~\ref{tab:different_graphs}. To predict the ridership at the next time interval (15 minutes), the baseline P-Net obtains a MAPE 19.04\% on SHMetro and 14.84\% on HZMetro, ranking last among all the variants. By aggregating the physical graph and any one of the proposed virtual graphs, the variants P+S-Net and P+C-Net achieve obvious performance improvements over all evaluation metrics. For instance, P+S-Net decreases the RMSE to from 50.45 to 47.38 on SHMetro and from 41.80 to 38.89 on HZMetro, while P+C-Net reduces the RMSE to 46.18 and 39.46. Moreover, we observe that the variant S+C-Net can also achieve very competitive performance, even though it does not contain the physical graph. On SHMetro dataset, S+C-Net obtains an RMSE 46.52, outperforming P-Net with a relative improvement of 7.8\%. On HZMetro dataset, S+C-Net also achieves a similar improvement by decreasing the RMSE to 39.92. These phenomenons indicate that the proposed virtual graphs are reasoning. Finally, the variant P+S+C-Net can obtain the best performance by incorporating the physical graph and all virtual graphs into networks. Specifically, P+S+C-Net gets the lowest RMSE (44.97 on SHMetro, 37.73 on HZMetro) and the lowest MAPE (16.83\% on SHMetro, 13.72\% on HZMetro). This significant improvement is mainly attributed to the enhanced spatial-temporal representation learned by the collaborative physical/virtual graph networks. These comparisons demonstrate the effectiveness of these tailor-designed graphs for the single time interval prediction.

Moreover, we find that these collaborative graphs are also effective for the ridership prediction of continuous time intervals. As shown in the bottom nine rows of Table~\ref{tab:different_graphs}, all variants suffer from performance degradation to some extent, as the number of time intervals increases from 2 to 4. For instance, the RMSE is rapidly increased to 73.06 on SHMetro and 56.32 on HZMetro, when the baseline P-Net is applied to forecast the ridership at the fourth time interval (60 minutes) of the future. By contrast, P+S-Net and P+C-Net achieve much lower RMSE (about 60 on SHMetro and 44 on HZMetro), since the proposed virtual graphs can prompt these variants to learn the complex flow patterns.
Incorporating all physical/virtual graphs, P+S+C-Net can further improve the performance with an RMSE 55.27 on SHMetro and 42.51 on HZMetro, which shows that these graphs are complementary.

\begin{table}[t]
  \caption{Effect of local feature and global feature. In our PVCGN, a Graph Convolution GRU is used to learn local feature and a Fully-Connected GRU is used to learn global feature.}
  \vspace{0mm}
\newcommand{\tabincell}[2]{\begin{tabular}{@{}#1@{}}#2\end{tabular}}
  \centering
  \resizebox{9cm}{!} {
    \begin{tabular}{c|c|c|c||c|c}
    \hline
    \multirow{2}{*}{Time} & \multirow{2}{*}{Metric} & \multicolumn{2}{|c||}{\textbf{SHMetro}} & \multicolumn{2}{c}{\textbf{HZMetro}} \\
    \cline{3-6}
     & & Local & Local + Global & Local & Local + Global\\
     \hline
     \hline
    \multirow{3}{*}{15 min}
     & \textit{RMSE}  &  45.64   & 44.97   & 38.46   & 37.76  \\
     & \textit{MAE~~} &  23.51   & 23.29   & 23.00   & 22.68  \\
     & \textit{MAPE}  &  17.23\% & 16.83\% & 13.86\% & 13.70\%\\
    \hline
    \multirow{3}{*}{30 min}
     & \textit{RMSE}  & 48.79   & 47.83   & 39.65   & 39.34  \\
     & \textit{MAE~~} & 24.48   & 24.16   & 23.78   & 23.33  \\
     & \textit{MAPE}  & 17.59\% & 17.23\% & 14.30\% & 13.81\%\\
    \hline
    \multirow{3}{*}{45 min}
     & \textit{RMSE}  & 52.70   & 52.02   & 41.45   & 40.95  \\
     & \textit{MAE~~} & 25.58   & 25.33   & 24.60   & 24.22  \\
     & \textit{MAPE}  & 18.16\% & 17.92\% & 14.88\% & 14.45\%\\
    \hline
    \multirow{3}{*}{60 min}
     & \textit{RMSE}  & 56.56   & 55.27   & 43.11   & 42.61  \\
     & \textit{MAE~~} & 26.50   & 26.29   & 25.36   & 24.93  \\
     & \textit{MAPE}  & 18.64\% & 18.69\% & 16.06\% & 15.49\%\\
    \hline
    \end{tabular}
  }
  \label{tab:local_global_feature}
  \vspace{0mm}
\end{table}

\subsubsection{\textbf{Influences of Local and Global Feature}}\label{sec:LG_effect}
As described in Section~\ref{sec:fusion}, a Graph Convolution Gated Recurrent Unit (GC-GRU) is developed for local feature learning, while a Fully-Connected Gated Recurrent Unit (FC-GRU) is applied to learn the global feature. In this section, we train two variants to explore the influence of each type of feature for metro ridership prediction. The first variant only contains GC-GRU, and the second variant consists of GC-GRU and FC-GRU. The results of these variants are summarized in Table~\ref{tab:local_global_feature}.
We can observe that the performance of the first variant is very competitive. For example, when predicting the ridership of the next 15 minutes, the first variant obtains an RMSE 45.64 on SHMetro and 38.46 on HZMetro. For the prediction of the fourth time interval, with a MAE 26.50 on SHMetro and 25.36 on HZMetro, this variant is slightly worse than the full model of PVCGN. This competitive performance is attributed to the fact that we can effectively learn the semantic local feature with the customized physical/virtual graphs. By fusing the local/global features of GC-GRU/FC-GRU, the second variant can boost the performance to a certain degree. For example, when predicting the ridership of the second time interval, the RMSE is decreased from 48.79 to 47.83 on SHMetro.
Through these experiments, we can conclude that the local feature plays a dominant role and the global feature provides ancillary information for ridership prediction.

\subsubsection{\textbf{Stability Verification}}
Following \cite{yao2019revisiting,song2020spatial}, we also examine the stability of the proposed PVCGN. Except for the formal model fully-evaluated in Section~\ref{sec:SOTA_Comp}, we implement another four models of PVCGN, because some random factors (e.g., parameter initialization, sample shuffle) may affect the final results. Due to the space limitation, the detailed performance of these extra models is shown in our supplementary material. The mean and standard deviation of all implemented models are summarized in Table~\ref{tab:stability}. We can observe that the mean performance is very close to that of the formal model and the deviation is very small on both benchmarks. Moreover, the worst model of our PVCGN still outperforms other compared methods. This experiment shows that PVCGN is stable.

\begin{table}[t]
  \caption{Mean and standard deviation of five implemented models of PVCGN on the whole testing sets.}
  \vspace{0mm}
\newcommand{\tabincell}[2]{\begin{tabular}{@{}#1@{}}#2\end{tabular}}
  \centering
    \begin{tabular}{c|c|c|c}
    \hline
    {Time} & {Metric} & {\textbf{SHMetro}} & {\textbf{HZMetro}} \\
     \hline
    \multirow{3}{*}{15 min}
     & \textit{RMSE}  &  45.09$\pm$0.22   & 37.83$\pm$0.24   \\
     & \textit{MAE~~} &  23.27$\pm$0.05   & 22.74$\pm$0.08   \\
     & \textit{MAPE}  &  17.05$\pm$0.38\% & 13.53$\pm$0.29\% \\
    \hline
    \multirow{3}{*}{30 min}
     & \textit{RMSE}  & 48.16$\pm$0.24   & 40.05$\pm$0.43   \\
     & \textit{MAE~~} & 24.19$\pm$0.06   & 23.66$\pm$0.20   \\
     & \textit{MAPE}  & 17.21$\pm$0.12\% & 13.87$\pm$0.28\% \\
    \hline
    \multirow{3}{*}{45 min}
     & \textit{RMSE}  & 51.90$\pm$0.20   & 41.82$\pm$0.56   \\
     & \textit{MAE~~} & 25.30$\pm$0.05   & 24.66$\pm$0.25   \\
     & \textit{MAPE}  & 18.04$\pm$0.23\% & 14.55$\pm$0.38\% \\
    \hline
    \multirow{3}{*}{60 min}
     & \textit{RMSE}  & 55.46$\pm$0.56   & 42.65$\pm$0.38   \\
     & \textit{MAE~~} & 26.28$\pm$0.09   & 25.13$\pm$0.16   \\
     & \textit{MAPE}  & 18.69$\pm$0.10\% & 15.46$\pm$0.65\% \\
    \hline
    \end{tabular}
  \label{tab:stability}
  \vspace{0mm}
\end{table}

\begin{table*}[t]
 \caption{Quantitative comparison for online origin-destination ridership prediction on the whole testing set of SHMetro.}
  \vspace{0mm}
\newcommand{\tabincell}[2]{\begin{tabular}{@{}#1@{}}#2\end{tabular}}
  \centering
    \begin{tabular}{c|c|c|c|c|c|c|c}
    \hline
    Time & Metric & {\textbf{HA}} & {\textbf{LSTM}} & {\textbf{GRU}} & {\textbf{DCRNN}} & {\textbf{GCRNN}} & {\textbf{PVCGN}} \\
    \hline
    \hline
    \multirow{3}{*}{15 min}
     & \textit{RMSE}  & 29.17   & 24.67   & 23.06   & 16.26   & 16.29   & 15.54   \\
     & \textit{MAE~~} & 5.76    & 5.47    & 5.30    & 4.69    & 4.69    & 4.54    \\
     & \textit{MAPE}  & 34.63\% & 25.50\% & 25.37\% & 24.56\% & 24.56\% & 23.63\% \\
    \hline
    \multirow{3}{*}{30 min}
     & \textit{RMSE}  & 29.1    & 24.49   & 23.44   & 17.88   & 17.66   & 16.51   \\
     & \textit{MAE~~} & 5.68    & 5.47    & 5.37    & 4.83    & 4.79    & 4.63    \\
     & \textit{MAPE}  & 34.57\% & 25.57\% & 25.54\% & 24.88\% & 24.78\% & 23.87\% \\
    \hline
    \multirow{3}{*}{45 min}
     & \textit{RMSE}  & 28.98   & 24.53   & 23.66   & 19.26   & 19.08   & 17.7    \\
     & \textit{MAE~~} & 5.59    & 5.47    & 5.42    & 4.93    & 4.91    & 4.77    \\
     & \textit{MAPE}  & 34.48\% & 25.55\% & 25.71\% & 25.28\% & 25.13\% & 24.20\% \\
    \hline
    \multirow{3}{*}{60 min}
     & \textit{RMSE}  & 28.75   & 24.71   & 23.75   & 20.88   & 20.6    & 18.61   \\
     & \textit{MAE~~} & 5.48    & 5.49    & 5.42    & 5.10    & 5.08    & 4.87    \\
     & \textit{MAPE}  & 34.40\% & 25.57\% & 25.67\% & 25.78\% & 25.66\% & 24.52\% \\
    \hline
    \end{tabular}
  \label{tab:onlineOD}
\end{table*}

\section{Apply to Online Origin-Destination Prediction}\label{sec:OD}
\begin{figure}[t]
    \centering
    \includegraphics[width=0.95\columnwidth]{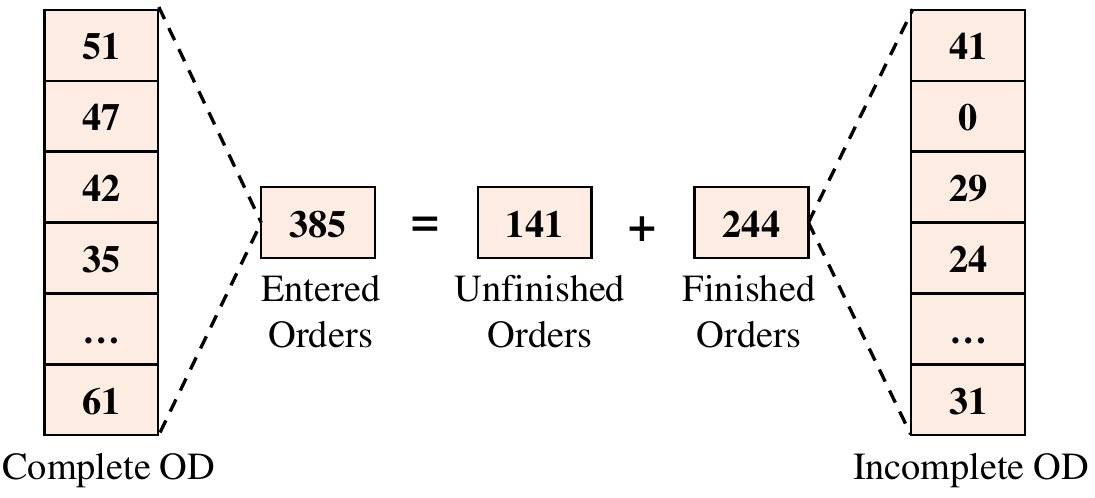}
    \vspace{-3mm}
    \caption{Illustration of the incomplete origin-destination (OD) distribution. In online metro systems, the complete OD distribution can not be obtained immediately. Suppose there were 385 passengers entered at the $i$-th station in the past 15 minutes and 244 of them have arrived at their destinations by now. The destinations of remaining passengers are unaware. In this case, we can only construct an incomplete OD vector from the finished orders.}
    \label{fig:onlineOD}
\end{figure}

In this section, we employ the proposed PVCGN to forecast the online metro origin-destination (OD) ridership on the SHMetro dataset. Compared with taxi OD demand prediction~\cite{liu2019contextualized}, metro OD ridership prediction is more challenging, because the complete OD distribution can not be obtained immediately in online metro systems~\cite{gong2020online}. For example, as shown in Fig.\ref{fig:onlineOD}, there were 385 passengers entered at the $i$-th station in the past 15 minutes and 244 of them have arrived at their destinations by now. The destinations of remaining passengers are unaware. Thus we can only construct an incomplete OD vector $\bm{X}_{t}^{I\_i}$ for station $i$ based on the finished orders.
Moreover, since the OD distribution is very sparse, we only consider the ridership from station $i$ to the top ten stations where its passengers are most likely to reach, as well as the total ridership to the remaining stations. Thus the length of OD vectors is 11. Specifically, $\bm{X}_{t}^{I\_i}(j)$ is the ridership to the $j$-th most relevant station, while $\bm{X}_{t}^{I\_i}(11)$ is the ridership to the remaining stations. for convenience, the incomplete OD ridership of all stations at time interval $t$ is denoted as $\bm{X}_{t}^I$=$(\bm{X}_t^{I\_1},\bm{X}_t^{I\_2},...,\bm{X}_t^{I\_N}) \in \mathbb{R}^{11 \times N}$.
Given a historical sequence of incomplete OD ridership, our goal is to forecast a future sequence of complete OD ridership:
\begin{equation}\nonumber
\begin{small}
\hat{\bm{X}}_{t+1}^{C},\hat{\bm{X}}_{t+2}^{C},..., \hat{\bm{X}}_{t+m}^{C}=\text{PVCGN}(\bm{X}_{t-n+1}^{I},\bm{X}_{t-n+2}^{I},...,\bm{X}_{t}^{I})
\end{small}
\end{equation}%
where $\hat{\bm{X}}_{t+1}^C$=$(\hat{\bm{X}}_{t+1}^{C\_1},\hat{\bm{X}}_{t+1}^{C\_2},...,\hat{\bm{X}}_{t+1}^{C\_N}) \in \mathbb{R}^{11 \times N}$ and $\hat{\bm{X}}_{t+1}^{C\_i}$ is the predicted complete OD ridership of station $i$. The sequences' lengths $n$ and $m$ are set to 4. As in the previous section, RMSE, MAE and MAPE are the evaluation metrics. When evaluating MAPE, we follow \cite{yao2018deep} to further filter some OD pairs with ground-truth ridership less than ten, since MAPE is sensitive to the small ridership and we also do not care about such low-ridership scenarios.

We compare our PVCGN with a baseline (i.e., Historical Average, HA) and four deep learning-based methods for online OD ridership prediction. As shown in Table~\ref{tab:onlineOD}, our PVCGN achieves superior performance at all time intervals and outperforms other methods with a substantial margin. Especially at the fourth interval, PVCGN decreases the MAPE to 24.52\% and has a relative improvement of 28.7\%, compared with the baseline HA. This is because our PVCGN can also learn the OD patterns effectively from our physical and virtual graphs. This experiment shows the universality of our PVCGN for online OD ridership prediction.

\section{Conclusion}\label{sec:conclusion}
In this work, we propose a unified Physical-Virtual Collaboration Graph Network to address the station-level metro ridership prediction. Unlike previous works that either ignored the topological information of a metro system or directly modeled on physical topology, we model the studied metro system as a physical graph and two virtual similarity/correlation graphs to fully capture the ridership evolution patterns. Specifically, the physical graph is built on the basis of the metro realistic topology. The similarity graph and correlation graph are constructed with virtual topologies under the guidance of the historical passenger flow similarity and correlation among different stations.
We incorporate these graphs into a Graph Convolution Gated Recurrent Unit (GC-GRU) to learn spatial-temporal representation and apply a Fully-Connected Gated Recurrent Unit (FC-GRU) to capture the global evolution tendency. Finally, these GRUs are utilized to develop a Seq2Seq model for forecasting the ridership of each station.
To verify the effectiveness of our method, we construct two real-world benchmarks with mass transaction records of Shanghai metro and Hangzhou metro and the extensive experiments on these benchmarks show the superiority of the proposed PVCGN.

In future works, we would pay more attention to the online origin-destination ridership prediction and several improvements should be considered. {\textbf{First}}, the data of unfinished orders can also provide some useful information and we attempt to estimate the potential OD distribution of unfinished orders. {\textbf{Second}}, the metro ridership evolves periodically. For instance, the ridership at 9:00 of every weekday is usually similar. Therefore, we should also utilize the periodic distribution of OD ridership to facilitate representation learning. {\textbf{Last but not least}}, some external factors (such as weather and holiday events) may greatly affect the ridership evolution and we should incorporate these factors to dynamically forecast the ridership.

\ifCLASSOPTIONcaptionsoff
  \newpage
\fi

\bibliographystyle{IEEEtran}
\bibliography{main}

\begin{IEEEbiography}[{\includegraphics[width=1in,height=1.25in,clip,keepaspectratio]{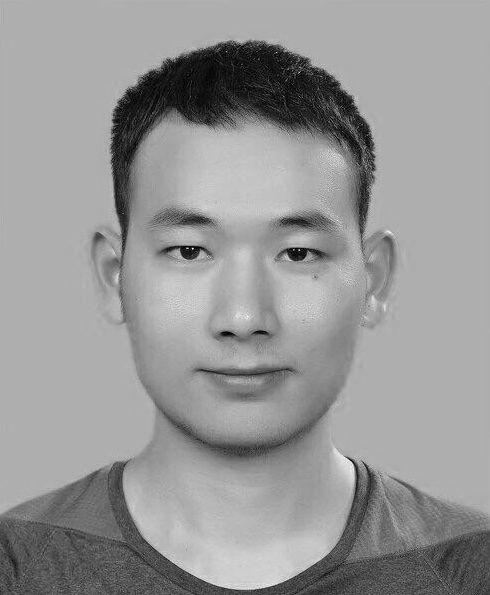}}]{Lingbo Liu} (Graduate Student Member, IEEE)
received the B.E. degree from the School of Software, Sun Yat-sen University, Guangzhou, China, in 2015, where he is currently
pursuing the Ph.D degree in computer science with the School of Data and Computer Science. From March 2018 to May 2019, he was a research assistant at the University of Sydney, Australia. His current research interests include machine learning and intelligent transportation systems. He has authorized and co-authorized on more than 10 papers in top-tier academic journals and conferences.
\end{IEEEbiography}

\begin{IEEEbiography}[{\includegraphics[width=1in,height=1.25in,clip,keepaspectratio]{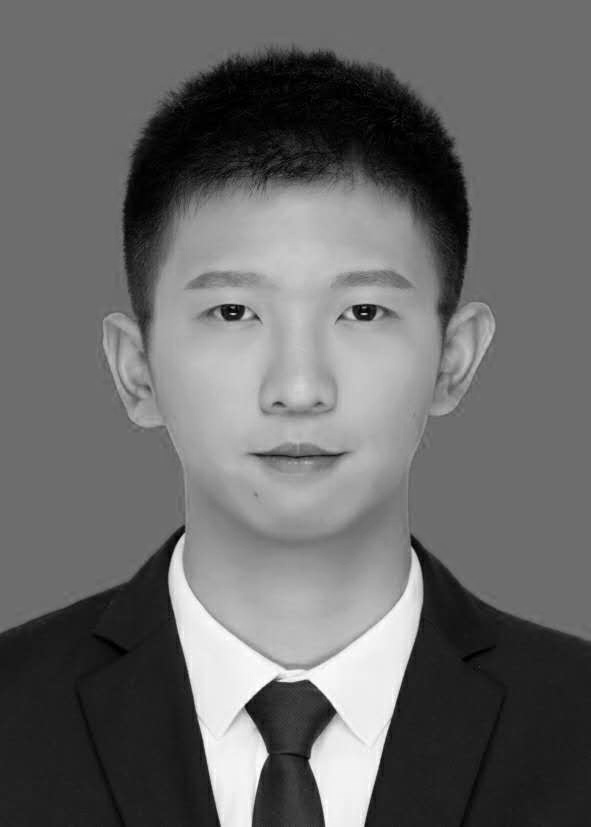}}]{Jingwen Chen}
received the B.E. degree from the School of Physics, Sun Yat-sen University, Guangzhou, China, in 2018, where he is currently pursuing the Master's degree in computer science with the School of Data and Computer Science. His current research interests include machine learning and data mining.
\end{IEEEbiography}

\begin{IEEEbiography}[{\includegraphics[width=1in,height=1.25in,clip,keepaspectratio]{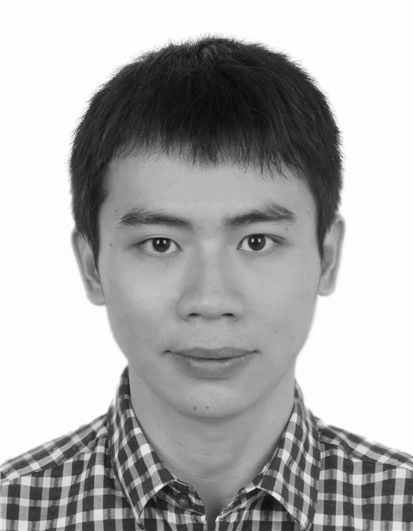}}]{Hefeng Wu} (Member, IEEE)
is currently a research associate professor in School of Data and Computer Science, Sun Yat-sen University. He is also a member of the National Engineering Research Center of Digital Life. He received the B.S. and Ph.D. degrees in computer science and technology from Sun Yat-sen University, China, in 2008 and 2013, respectively. From 2014 to 2018, he was an Assistant Professor with the School of Information Science and Technology, Guangdong University of Foreign Studies, Guangzhou, China. His research interests include machine learning and computer vision.
\end{IEEEbiography}

\begin{IEEEbiography}[{\includegraphics[width=1in,height=1.25in,clip,keepaspectratio]{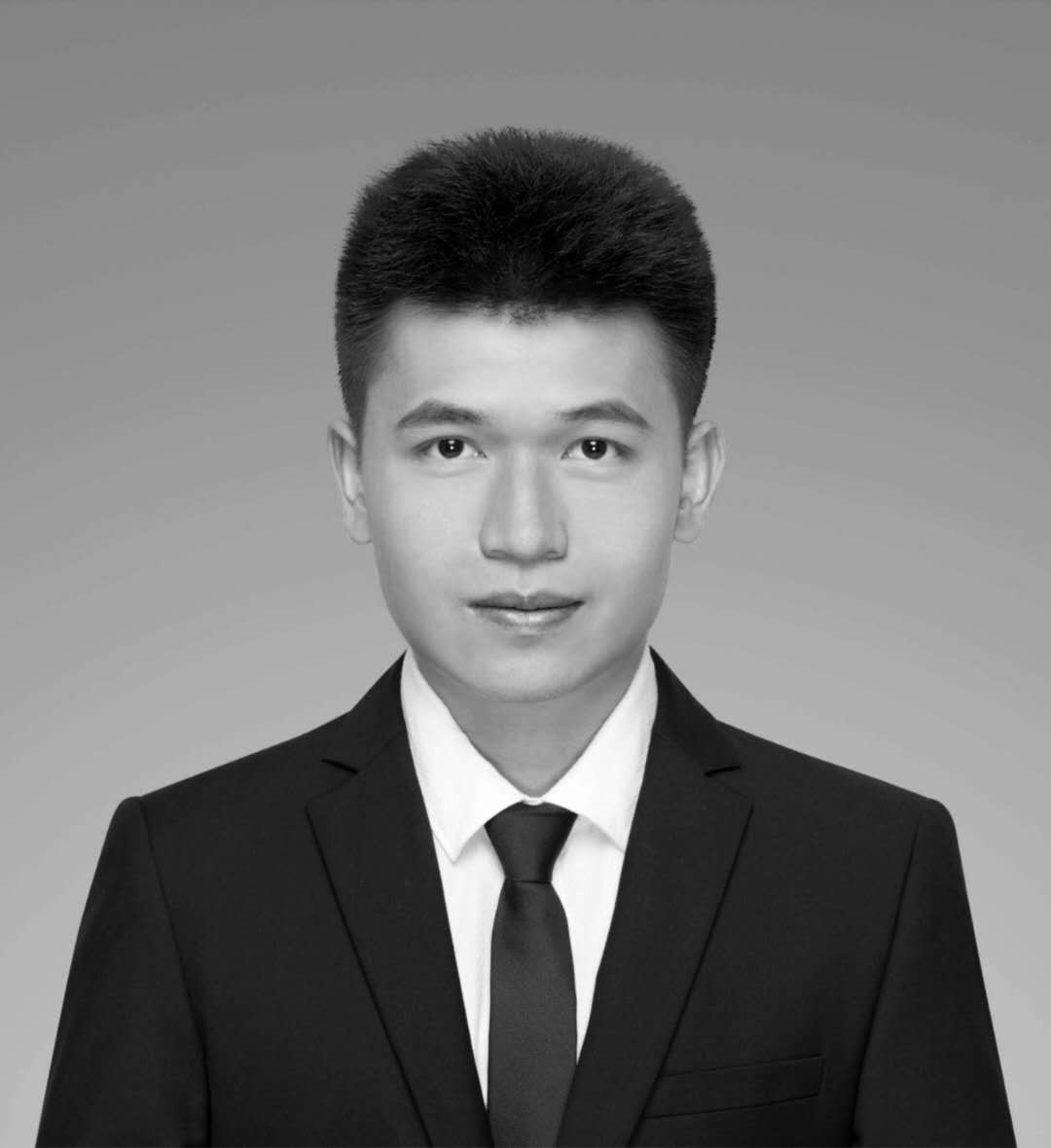}}]{Jiajie Zhen}
received the B.E. degree from the School of Mathematics, Sun Yat-sen University, Guangzhou, China, in 2018, where he is currently pursuing the Master's degree in computer science with the School of Data and Computer Science. His current research interests include data mining and intelligent transportation systems.
\end{IEEEbiography}

\begin{IEEEbiography}[{\includegraphics[width=1in,height=1.25in,clip,keepaspectratio]{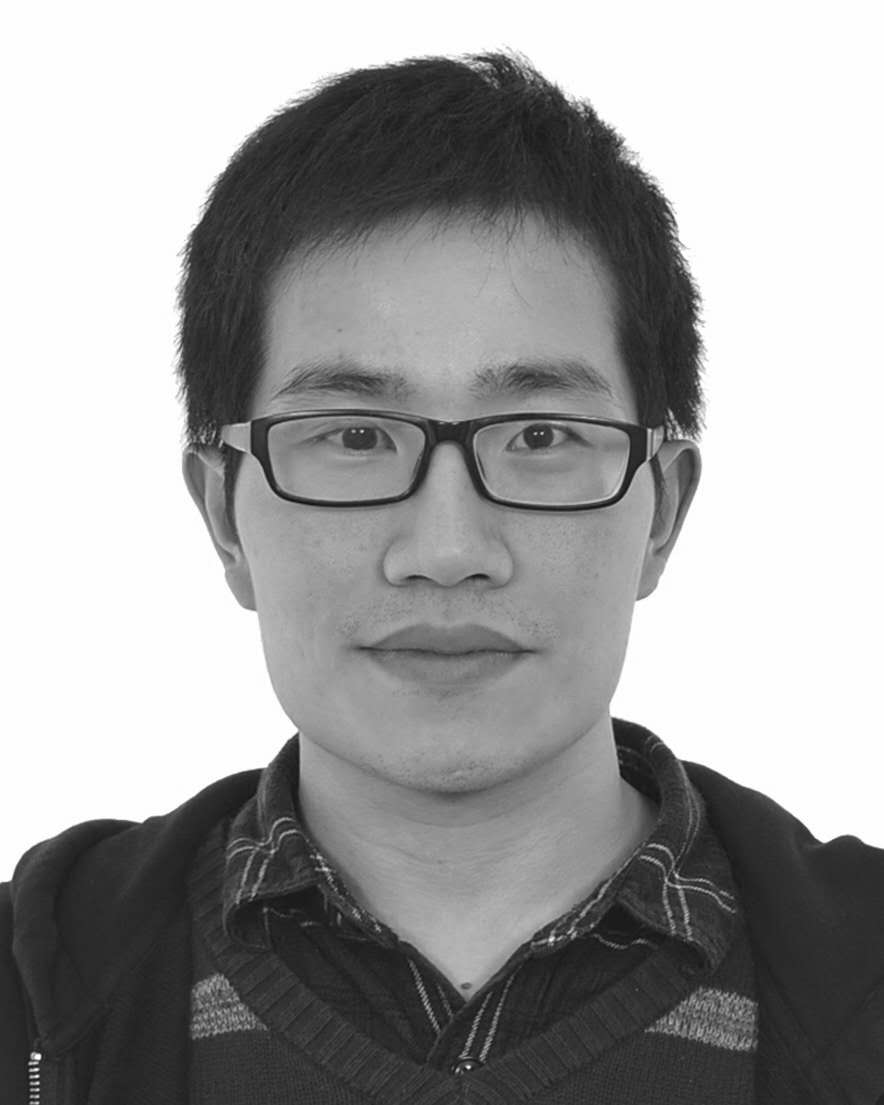}}]{Guanbin Li} (Member, IEEE) is currently an associate professor in School of Data and Computer Science, Sun Yat-sen University. He received his PhD degree from the University of Hong Kong in 2016. His current research interests include computer vision, image processing, and deep learning. He is a recipient of ICCV 2019 Best Paper Nomination Award. He has authorized and co-authorized on more than 60 papers in top-tier academic journals and conferences. He serves as an associate editor for journal of The Visual Computer, an area chair for the conference of VISAPP. He has been serving as a reviewer for numerous academic journals and conferences such as TPAMI, IJCV, TIP, TMM, TCyb, CVPR, ICCV, ECCV and NeurIPS.
\end{IEEEbiography}

\begin{IEEEbiography}[{\includegraphics[width=1in,height=1.25in,clip,keepaspectratio]{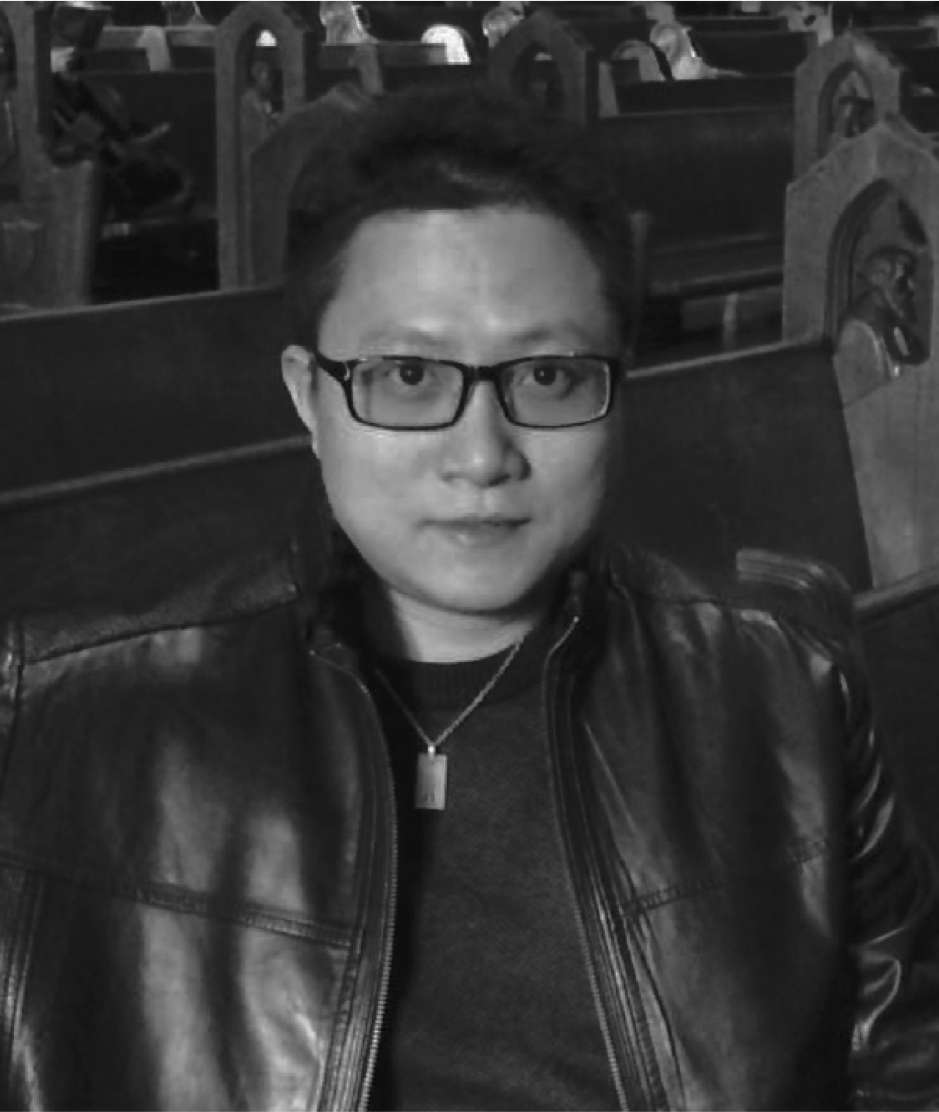}}]{Liang Lin} (Senior Member, IEEE) is a full Professor of Sun Yat-sen University. He is the Excellent Young Scientist of the National Natural Science Foundation of China. From 2008 to 2010, he was a Post-Doctoral Fellow at University of California, Los Angeles. From 2014 to 2015, as a senior visiting scholar, he was with The Hong Kong Polytechnic University and The Chinese University of Hong Kong. From 2017 to 2018, he leaded the SenseTime R\&D teams to develop cutting-edges and deliverable solutions on computer vision, data analysis and mining, and intelligent robotic systems. He has authorized and co-authorized on more than 100 papers in top-tier academic journals and conferences. He has been serving as an associate editor of IEEE Trans. Neural Networks and Learning Systems, IEEE Trans. Human-Machine Systems, The Visual Computer and Neurocomputing. He served as Area/Session Chairs for numerous conferences such as ICME, ACCV, ICMR. He was the recipient of Best Paper Nomination Award in ICCV 2019, Best Paper Runners-Up Award in ACM NPAR 2010, Google Faculty Award in 2012, Best Paper Diamond Award in IEEE ICME 2017, and Hong Kong Scholars Award in 2014. He is a Fellow of IET.
\end{IEEEbiography}

\end{document}